\pdfoutput=1                            
\documentclass{article}
\usepackage{iclr2027_conference,times}

\usepackage{amsmath,amsfonts,bm}

\def\eqref#1{equation~\ref{#1}}

\def\1{\bm{1}}

\def\vmu{{\bm{\mu}}}

\def\vc{{\bm{c}}}

\def\vr{{\bm{r}}}
\def\vs{{\bm{s}}}

\def\vv{{\bm{v}}}

\def\vz{{\bm{z}}}

\def\mI{{\bm{I}}}

\DeclareMathAlphabet{\mathsfit}{\encodingdefault}{\sfdefault}{m}{sl}
\SetMathAlphabet{\mathsfit}{bold}{\encodingdefault}{\sfdefault}{bx}{n}

\newcommand{\R}{\mathbb{R}}

\usepackage{hyperref}
\usepackage{url}
\usepackage{amsmath,amssymb}
\usepackage{booktabs}
\usepackage{multirow}
\usepackage{graphicx}
\usepackage{xcolor}
\usepackage{float}
\usepackage{array}                      
\usepackage{placeins}                   
\usepackage{tikz}
\usepackage{titletoc}

\usepackage{todonotes}
\usetikzlibrary{arrows.meta,positioning,fit,calc}
\usepackage{tcolorbox}                  
\tcbuselibrary{breakable}
\newtcolorbox{samplebox}[1]{breakable, colback=white, colframe=black!55, colbacktitle=black!7, coltitle=black,
  fonttitle=\bfseries\small, title={#1}, arc=1.5mm, boxrule=0.5pt, left=2mm, right=2mm, top=1.2mm, bottom=1.2mm,
  before skip=8pt, after skip=8pt}

\newcommand{\method}{PSF}               
\newcommand{\ours}{\method{}}
\newcommand{\genppl}{Gen.\ PPL}
\newcommand{\iso}{@5.16}                
\newcommand{\vsigma}{\boldsymbol{\sigma}}
\newcommand{\vepsilon}{\boldsymbol{\epsilon}}

\title{Rethinking the Roles of Diffusion and Decoding: Predictive-State Flows for Parallel Language Generation}

\author{Jinjin He$^{1*}$, Xinsong Feng$^{1*}$, Lucheng Fu$^{1}$, Sinan Wang$^{1}$, Bo Zhu$^{1}$, Jiayun Wang$^{1}$ \\
$^{1}$Georgia Institute of Technology \\
{\small\texttt{\{jhe433,xsfeng,luchengfu,swang3081,bo.zhu,pjwang\}@gatech.edu}}
}

\iclrfinalcopy 
\begin{document}
\maketitle
\lhead{Preprint}                        
{\renewcommand{\thefootnote}{\fnsymbol{footnote}}\footnotetext[1]{Equal contribution.}}

\begin{abstract}

In continuous diffusion language models (DLMs), latents generated by the diffusion prior are mapped almost deterministically to tokens, either by a decoder or through a token embedding matrix.
Unlike in image generation, where visual details can vary continuously, the diffusion prior must capture fine-grained discrete token choices in continuous space, which can make the joint latent distribution difficult to model.
We argue that generative roles should instead be divided: the diffusion prior should generate high-level \emph{predictive information}, while the decoder should resolve the remaining token-level choices.
Based on this view, we introduce \emph{Predictive-State Flows} (\method{}), which partitions a sequence into blocks and associates each block with a predictive state computed solely from the preceding context.
A rectified-flow prior generates the sequence of predictive states, after which a block-causal decoder generates all blocks in parallel, autoregressively only within each block.
This design reduces diffusion from $L$ token-level states to $K \ll L$ predictive states and reduces token-level sequential depth to $L/K$.
On unconditional OpenWebText generation, a 197M-parameter model with $K=8$ achieves a generative perplexity of 22.8 at entropy 5.16 using 0.31 TFLOPs per 1024-token document, compared with 24.0 and 7.2 TFLOPs for ELF-B with 32 sampling steps.
The corresponding measured operating points yield MAUVE@512 of 0.93 and 0.73, respectively.
Controlled comparisons further show that content states, which additionally encode their own blocks, yield 8--16\% higher generative perplexity than predictive states, and that replacing sampled predictive-state trajectories with a shared mean lowers MAUVE@512 by 0.11.
These results suggest that latent design for continuous DLMs should consider not only how compactly text is represented, but also how generative roles are divided between the prior and the decoder.
\end{abstract}

\section{Introduction}

Diffusion language models (DLMs) generate text by iteratively refining multiple token positions in parallel rather than decoding strictly from left to right~\citep{li2022diffusion,sahoo2024simple,nie2026large}.
Continuous DLMs perform this refinement over continuous states that are subsequently decoded into text~\citep{gulrajani2023likelihood,hu2026elf}.
In many existing formulations, the generated states are mapped nearly deterministically to tokens~\citep{dieleman2022continuous,hu2024flow}.
This assigns most generative roles to the diffusion prior: it must specify fine-grained token choices, while the decoder largely recovers choices already encoded in those states.
This raises a fundamental question: \emph{which information should be generated by the diffusion prior, and which decisions should be left to the decoder?}


This division of roles arises in part from how continuous text representations are commonly learned.
An encoder maps an observed token sequence to latents, and a decoder is trained to reconstruct the same sequence~\citep{lovelace2023latent,zhang2023planner}.
Such an objective encourages the latents to preserve the specific token choices that distinguish the observed text from other plausible realizations, leaving the prior responsible for generating those choices.
Unlike in image generation, where perceptual compression can discard visually negligible details and represent the remaining variation continuously~\citep{rombach2022high,esser2024scaling}, reconstruction-based text latents tend to preserve fine-grained discrete token distinctions.
The diffusion prior must therefore model these distinctions and their dependencies in continuous space, making the joint latent distribution difficult to learn.
We therefore propose a different division of generative roles: the diffusion prior generates \emph{predictive information} that guides subsequent content, while a decoder resolves the remaining local token choices.



\begin{figure}[t]
\centering
\begingroup%
\definecolor{ovPredict}{HTML}{C9D7EC}%
\definecolor{ovLocal}{HTML}{F0DFC8}%
\definecolor{ovBlue}{HTML}{476695}%
\definecolor{ovOchre}{HTML}{986A3A}%
\resizebox{\linewidth}{!}{%
\begin{tikzpicture}[
  x=1cm,y=-1cm,
  font=\sffamily\fontsize{8.5}{10}\selectfont,
  every node/.style={inner sep=0pt,outer sep=0pt},
  heading/.style={text=black!60,font=\sffamily\fontsize{8}{9}\selectfont},
  rowtitle/.style={anchor=west,font=\sffamily\bfseries\fontsize{9}{10.5}\selectfont},
  note/.style={font=\sffamily\fontsize{7.7}{9}\selectfont,text=black!65},
  formula/.style={font=\fontsize{10}{11}\selectfont},
  legend/.style={anchor=west,align=left,font=\sffamily\fontsize{7.7}{8.3}\selectfont},
  outline/.style={draw=black!70,line width=.45pt,rounded corners=2pt},
  cell/.style={line width=.35pt},
  arrow/.style={-{Latex[length=1.4mm,width=.9mm]},draw=black!80,line width=.5pt}
]
\path[use as bounding box] (0,0) rectangle (14,5.48);

\node[heading] at (1.55,.22) {Encoder input};
\node[heading] at (5.78,.22) {State modeled by the prior};
\node[heading] at (9.38,.22) {Decoder};
\node[heading] at (12.45,.22) {Text};
\path[fill=black!4,rounded corners=4pt] (0,.55) rectangle (14,2.95);
\path[fill=black!4,rounded corners=4pt] (0,3.06) rectangle (14,5.48);

\node[rowtitle] at (.25,.89) {(a) Content latents};
\node at (1.55,1.49) {current block $x^{(k)}$};
\foreach \word/\xx in {the/.84,fox/1.55,ran/2.26} {
  \node[outline,fill=white,minimum width=.64cm,minimum height=.53cm]
    at (\xx,2.03) {\word};
}
\draw[arrow] (2.88,2.03) -- (3.80,2.03);
\node[note] at (3.32,1.75) {encode};
\node[formula,anchor=west] at (4.22,1.17) {$\vz_k=E(x^{(k)})$};
\path[outline,fill=white] (4.10,1.50) rectangle (4.90,2.54);
\foreach \yy in {1.64,1.87,2.29} {
  \foreach \j in {0,1,2,3} {
    \ifnum\j<2
      \path[cell,draw=ovBlue,fill=ovPredict]
        ({4.22+.14*\j},\yy) rectangle ({4.36+.14*\j},{\yy+.15});
    \else
      \path[cell,draw=ovOchre,fill=ovLocal]
        ({4.22+.14*\j},\yy) rectangle ({4.36+.14*\j},{\yy+.15});
    \fi
  }
}
\node[font=\fontsize{8}{8}\selectfont] at (4.50,2.13) {$\cdots$};
\path[cell,draw=ovBlue,fill=ovPredict] (5.44,1.65) rectangle (5.58,1.79);
\node[legend,text=ovBlue] at (5.76,1.72) {predictive\\information};
\path[cell,draw=ovOchre,fill=ovLocal] (5.44,2.29) rectangle (5.58,2.43);
\node[legend,text=ovOchre] at (5.76,2.36) {specific token\\choices};
\node[note] at (5.78,2.78) {prior models both};
\draw[arrow] (7.82,2.03) -- (8.22,2.03);
\node[outline,fill=white,minimum width=2.04cm,minimum height=.78cm]
  at (9.38,2.03) {decoder};
\draw[arrow] (10.62,2.03) -- (11.13,2.03);
\node[note] at (12.45,1.49) {specific block};
\node[text=ovOchre] at (12.45,2.03) {the fox ran};

\node[rowtitle] at (.25,3.38) {(b) Predictive states (ours)};
\node at (1.55,3.96) {prefix $x^{(<k)}$};
\node[outline,fill=white,minimum width=2.20cm,minimum height=.61cm]
  at (1.55,4.52) {\ldots\ the fox};
\node[note] at (1.55,5.16) {block $k$ is unseen};
\draw[arrow] (2.88,4.52) -- (3.80,4.52);
\node[note] at (3.32,4.24) {encode};
\node[formula,anchor=west] at (4.22,3.76) {$\vs_k=E_\psi(x^{(<k)})$};
\path[outline,fill=white] (4.10,4.00) rectangle (4.90,5.04);
\foreach \yy in {4.14,4.37,4.79} {
  \foreach \j in {0,1,2,3} {
    \path[cell,draw=ovBlue,fill=ovPredict]
      ({4.22+.14*\j},\yy) rectangle ({4.36+.14*\j},{\yy+.15});
  }
}
\node[font=\fontsize{8}{8}\selectfont] at (4.50,4.63) {$\cdots$};
\path[cell,draw=ovBlue,fill=ovPredict] (5.44,4.45) rectangle (5.58,4.59);
\node[legend,text=ovBlue] at (5.76,4.52) {predictive\\context};
\node[note] at (5.78,5.26) {prior models $\vs_{1:K}$ jointly};
\draw[arrow] (7.82,4.52) -- (8.22,4.52);
\node[text=ovBlue,font=\fontsize{8.5}{9.5}\selectfont]
  at (8.02,4.12) {$\vs_{\le k}$};
\path[outline,fill=white] (8.36,4.05) rectangle (10.40,4.99);
\node at (9.38,4.30) {block decoder};
\node[fill=ovLocal,rounded corners=1pt,minimum width=1.76cm,minimum height=.32cm]
  at (9.38,4.73) {local details};
\draw[arrow] (10.62,4.52) -- (11.13,4.52);
\node[note] at (12.45,3.96) {realized block};
\node[text=ovOchre,align=center] at (12.45,4.52) {ran across\\[2pt]the field};
\end{tikzpicture}%
}%
\endgroup%
%
\caption{\textbf{Content latents and predictive states.} (a) A content latent encodes the current block, carrying predictive information and specific token choices. (b) In \ours{}, $\vs_k$ encodes only the prefix before block $k$; the prior models predictive states jointly and the decoder predicts local realizations. Small cells schematically indicate predictive information (blue) and token choices (tan). Landing targets and optional realization codes are omitted; Appendix~\ref{app:architecture} gives details.}
\label{fig:overview}
\end{figure}
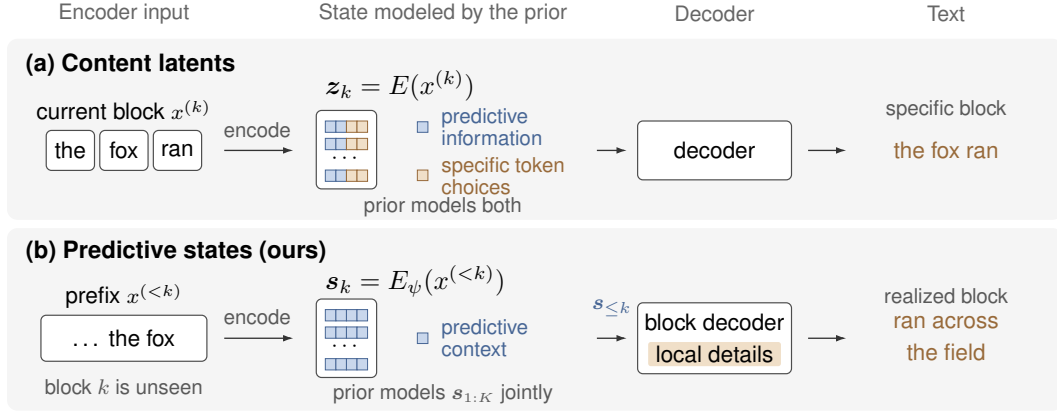

We instantiate this principle using \emph{predictive states}.
Given a sequence $x_{1:L} \in \mathcal{V}^L$ divided into $K$ consecutive blocks, we associate each block $k$ with a continuous state
\begin{equation}
\vs_k = E_\psi\left(x^{(<k)}\right),
\end{equation}
which summarizes the context preceding block $k$ without observing the block itself.
The state therefore provides information for predicting the upcoming block, rather than directly encoding its specific token choices.
We use a rectified-flow prior~\citep{liu2022flow} to generate the predictive-state trajectory $\vs_{1:K}$.
Conditioned on these states, a block-causal decoder predicts all $K$ blocks in parallel, with autoregressive decoding restricted to within each block.
We call the resulting framework \emph{Predictive-State Flow} (\method{}). 
Figure~\ref{fig:overview} contrasts content latents with predictive states; Appendix~\ref{app:architecture} provides diagrams of the encoder, flow prior, and parallel block decoder.

However, this division of roles introduces a consistency challenge.
Because the full predictive-state trajectory is generated before token decoding, independently decoded blocks may become inconsistent at their boundaries.
We coordinate adjacent blocks through a one-token \emph{landing target}: after the first parallel decoding step, block $k$ can condition on the first generated token of block $k+1$, providing a lightweight constraint on their transition.
We further adapt the decoder to rollout states produced by the learned prior, reducing the train--generation mismatch.
This division also supports parallel generation: the prior operates on $K \ll L$ predictive states, while the decoder generates the corresponding blocks concurrently.
With autoregression restricted to within each block, the decoder requires only $L/K$ sequential token-generation steps instead of $L$.

Our experiments directly evaluate this division of generative roles.
On unconditional OpenWebText generation, a 197M-parameter \method{} model with $K=8$ achieves a generative perplexity of 22.8 at entropy 5.16 while requiring only 0.31\,TFLOPs per 1024-token sequence, compared with 24.0 and 7.2\,TFLOPs for ELF-B.
In a controlled comparison, content states that also encode their own block are 8--16\% worse than predictive states.
Under nucleus sampling, PSF achieves a generative perplexity of 19.4 when the prior generates only predictive states, compared with 21.2 when it additionally generates latent variables carrying information about each block’s token realization. 
This supports leaving local token choices to the decoder at long block lengths.

Further reducing $K$ from 64 to 8 also improves perplexity from 39.2 to 22.8 while lowering the computational cost of the diffusion prior. 
This reflect a trade-off between the latent-generation difficulty and the amount of local autoregressive realization.
Same-hardware measurements further show that lower FLOPs do not translate directly into wall-clock speedups, as latency and throughput vary with both block length and batch size.

Our contributions are summarized as follows.
First, we introduce \emph{predictive states}, which allow the diffusion prior to model high-level predictive information while leaving token-level realization to the decoder.
Second, we develop \method{}, which generates a short predictive-state trajectory with rectified flow and decodes the corresponding blocks in parallel, using one-token landing targets to coordinate adjacent blocks and local autoregression within each block to limit sequential decoding depth.
Third, we empirically study this division of generative roles, showing that predictive states outperform content states and that strong generation quality at long block lengths can be achieved without requiring the prior to encode block-specific token information.

\section{Predictive-State Flow}
\label{sec:method}

We begin by separating the generation of predictive information from token-level realization.
\method{} uses a rectified-flow prior to generate a trajectory of \emph{predictive states}, while a predictive block decoder generates the corresponding token-level realizations.
Each predictive state summarizes the context preceding a block without encoding the block itself, allowing the prior to model predictive information while leaving local token choices to the block decoder.
The block decoder realizes all blocks in parallel, while a one-token \emph{landing target} couples adjacent branches to improve consistency across block boundaries.

\subsection{Predictive States}
\label{sec:encoder}

We divide a sequence $x_{1:L}\in \mathcal{V}^L$ into $K$ consecutive blocks $x^{(1)},\dots,x^{(K)}$ of length $\ell=L/K$ and associate each block $k$ with a predictive state $\vs_k\in\mathbb{R}^d$ that summarizes the context preceding that block.
To ensure that each state contains only information available from the corresponding prefix, we use a block-causal encoder in which representations in block $j$ depend only on tokens in blocks $1,\dots,j$.
We mean-pool the outputs of block $j$ to obtain a context state
\begin{equation}
\vc_j=f_\psi\!\left(x^{(\le j)}\right).
\end{equation}

We then shift these states by one block so that the state associated with block $k$ depends only on its preceding context:
\begin{equation}
\vs_1=\vs_{\mathrm{bos}},
\qquad
\vs_k=\vc_{k-1}=f_\psi\!\left(x^{(<k)}\right),
\quad k=2,\dots,K,
\end{equation}
where $\vs_{\mathrm{bos}}$ is a learned state representing the empty prefix.
By construction, $\vs_k$ depends only on $x^{(<k)}$, unlike content latents that encode a particular realization of their own block~\citep{lovelace2023latent,zhang2023planner}.
We refer to the predictive-state trajectory $\mathbf{S}=(\vs_1,\dots,\vs_K)$, optionally together with the realization codes introduced in Section~\ref{sec:training}, as the document \emph{plan}.
The diffusion prior models this plan, while the block decoder resolves the token-level realization of each block.

\subsection{Parallel Block Decoder with Landing Targets}
\label{sec:writer}

Given the predictive states, the block decoder generates the $K$ blocks in parallel, with causal self-attention within each block.
Block $k$ cross-attends to $\vs_{\le k}$ under a causal mask over the block index.

Because predictive states do not specify a unique local realization, independently generated blocks can each be plausible while remaining inconsistent across block boundaries.
We address this issue with a one-token \emph{landing target}.
The first decoding step generates the first token of every block in parallel.
From the second step onward, block $k$ additionally conditions on the first token of block $k+1$.
During training, this target is the ground-truth token, whereas during generation it is the token produced at the first decoding step.
Let $\tau_k=x^{(k+1)}_1$ for $k<K$, and let $\tau_K$ denote the learned ``no-target'' embedding supplied to the final block.
The block decoder factorizes as
\begin{equation}
p_\theta\!\left(x^{(1:K)}\mid\boldsymbol{s}_{1:K}\right)
=
\prod_{k=1}^{K}
\left[
p_\theta\!\left(x^{(k)}_1\mid\boldsymbol{s}_{\le k}\right)
\prod_{i=2}^{\ell}
p_\theta\!\left(
x^{(k)}_i
\mid
x^{(k)}_{<i},
\boldsymbol{s}_{\le k},
\tau_k
\right)
\right].
\label{eq:writer}
\end{equation}

After the first decoding step, all landing targets are available, and the $K$ branches proceed in parallel for the remaining steps.
The decoder therefore requires only $\ell$ sequential token-generation steps, rather than $L$ for standard autoregressive decoding.

\subsection{Flow Prior over Predictive States}
\label{sec:prior}

We model the joint distribution of predictive-state trajectories with a rectified-flow prior~\citep{liu2022flow,lipman2022flow}.
After training the encoder and decoder, we freeze both and use the encoder to map each training sequence to a trajectory $\mathbf{S}=(\vs_1,\dots,\vs_K)$.
We standardize the predictive states using training-set statistics before feeding them to the prior.

Given $\vepsilon\sim\mathcal{N}(\mathbf{0},\mI)$, we construct $\mathbf{S}_t=t\mathbf{S}+(1-t)\vepsilon$.
We use an $x$-prediction parameterization, where the prior predicts the clean trajectory $\hat{\mathbf{S}}_\phi(\mathbf{S}_t,t)$.
The corresponding velocity prediction is
\begin{equation}
\hat{\vv}_\phi(\mathbf{S}_t,t)
=
\frac{\hat{\mathbf{S}}_\phi(\mathbf{S}_t,t)-\mathbf{S}_t}{1-t},
\end{equation}
and we optimize the rectified-flow objective
\begin{equation}
\mathcal{L}_{\mathrm{flow}}
=
\mathbb{E}_{\mathbf{S},\vepsilon,t}
\left[
\left\|
\hat{\vv}_\phi(\mathbf{S}_t,t)
-
(\mathbf{S}-\vepsilon)
\right\|_2^2
\right].
\end{equation}

Following ELF, we sample $t$ from a logit-normal distribution with mean $-1.5$ and standard deviation $0.8$, use self-conditioning with probability $0.5$, and apply classifier-free guidance on the self-conditioning input through guidance-scale tokens.
Guidance scales are sampled log-uniformly during training and selected at sampling time.
At generation time, we sample $\mathbf{S}_0\sim\mathcal{N}(\mathbf{0},\mI)$ and integrate the induced velocity field from $t=0$ to $t=1$ using 50 Euler steps.

\subsection{Training}
\label{sec:training}

Training of \method{} proceeds in three stages.
We first jointly train the encoder and block decoder for reconstruction.
We then freeze the encoder trunk and fine-tune the decoder to improve robustness to generated predictive states, introduce block-specific realization codes, and adapt the model across different block lengths.
Finally, we freeze the encoder and decoder and train the flow prior over the resulting plans.
Model-specific training schedules and hyperparameters are provided in Appendix~\ref{app:hparams}.

\textbf{Stage I: reconstruction training.}
We first jointly train the encoder and block decoder using token-level cross-entropy, adding Gaussian noise to the predictive states read by the decoder.
For block $k$, the decoder conditions on the predictive states $\vs_{\le k}$, which summarize only the preceding blocks.

\textbf{Stage II: decoder fine-tuning.}
We next freeze the encoder trunk and further train the block decoder in three steps: adapting it to rollout states, introducing block-specific realization codes, and adapting it across different block lengths.

\emph{Training on generated states.}
During initial training, the block decoder conditions on predictive states encoded from ground-truth sequences.
At generation time, however, it conditions on states produced by the flow prior rather than states directly encoded from ground-truth text.
To reduce this mismatch, we generate a no-gradient rollout conditioned on the ground-truth predictive states and re-encode the resulting sequence.
For each block independently, we replace the original predictive state with its re-encoded counterpart with probability $0.5$, while keeping the original token targets unchanged.
The rollout uses temperature $0.65$ and nucleus sampling with $p=0.95$.

We provide a binary \emph{source label} $\lambda_k$ for each predictive state, where $\lambda_k=1$ denotes an original encoder state and $\lambda_k=0$ denotes a rollout state.
A zero-initialized embedding of $\lambda_k$ is added to $\vs_k$.
At generation time, we set $\lambda_k=0$ for all blocks.

\emph{Realization codes.}
\label{sec:codes}
The predictive state $\vs_k=\vc_{k-1}$ is computed before block $k$ is observed and therefore contains only information from the preceding blocks.
We next augment the decoder with a low-dimensional realization code $\vr_k\in\mathbb{R}^{64}$ from the unshifted encoder state $\vc_k=f_\psi(x^{(\le k)})$, which additionally contains information from block $k$.
A newly added linear head predicts the mean and diagonal variance of a Gaussian posterior,
\begin{equation}
q_\psi(\vr_k\mid\vc_k)
=
\mathcal{N}\!\left(
\vmu_\psi(\vc_k),
\operatorname{diag}\!\left(\vsigma_\psi^2(\vc_k)\right)
\right).
\end{equation}
The block decoder conditions on both $\vr_k$ and the next-block code $\vr_{k+1}$, sampled from their posteriors during training.
During this stage, the encoder trunk remains frozen, while the decoder, code head, and newly introduced code-conditioning layers are trained jointly.
We regularize the posterior toward $\mathcal{N}(\mathbf{0},\mI)$ with a KL budget of $\ell$ nats per block that limits the amount of block-specific information carried by $\vr_k$.
Further details of the posterior and conditioning architecture are provided in Appendix~\ref{app:codes}.

\emph{Training across block lengths.}
Starting from $K=64$, we successively fine-tune the decoder at $K=32$, $16$, and $8$, while keeping the encoder trunk frozen.

\textbf{Stage III: flow-prior training.}
Finally, we freeze the encoder and block decoder and train the flow prior described in Section~\ref{sec:prior}.
For each training sequence, we encode the predictive states $\vs_k$ and use the posterior mean $\vmu_k$ as the target realization code.
The prior is trained to jointly generate these representations using the flow-matching objective defined in the previous section.
We train a separate prior for each value of $K$, warm-started from the previous block length.

For $K=8$, we additionally train a \emph{states-only prior} that generates only the predictive states, using zero code targets.
We retain the code-trained block decoder and fix all its realization-code inputs to $\mathbf{0}$.
At inference time, we use generated codes for $K\ge16$ and the states-only prior for $K=8$ (Section~\ref{sec:ablations}).

\subsection{Generation and Refinement}
\label{sec:generation}

For generation, we first sample the predictive-state trajectory from the flow prior, together with realization codes when they are used.
The block decoder then generates all $K$ blocks in parallel over $\ell=L/K$ autoregressive steps.
The first decoding step produces the landing targets, after which each branch conditions on the first token of the following block and continues independently.
We set $\lambda_k=0$ for all blocks during generation.
The resulting serial depth is $50+\ell$, compared with $L$ steps for standard autoregressive generation.

Optionally, we refine the generated document by re-encoding it and running the block decoder again.
When realization codes are used, they are recomputed from the re-encoded sequence.
Because refinement conditions on encoder-produced states, we set $\lambda_k=1$.
Each refinement round requires one encoder pass followed by another $\ell$-step parallel decode.

\section{Experiments}
\label{sec:experiments}

\textbf{Dataset and evaluation.}
We train and evaluate on OpenWebText~\citep{Gokaslan2019OpenWeb} with the T5 tokenizer ($V{=}32{,}100$)~\citep{raffel2020exploring}, $L{=}1024$ tokens per document and 10,000 held-out documents, as in ELF. Each setting generates 1000 documents; every document is scored up to its first end-of-text token by a frozen GPT-2 Large~\citep{radford2019language}, giving the corpus-level generative perplexity (\genppl{}). We pair it with the \emph{entropy}, the unigram entropy of GPT-2 tokens averaged over documents. To compare at matched entropy, we sweep the block-decoder temperature $t$ at nucleus 0.95 (the \emph{plain sampler}) or the nucleus threshold $p$ at $t{=}1$~\citep{holtzman2019curious} and interpolate log-linearly to 5.16, the rounded entropy of ELF-B's 32-step samples (marked ``\iso''). MAUVE uses the first 512 judge tokens (Appendix~\ref{app:metrics}). Sampling-seed standard deviations are 0.04--0.4 perplexity points (Appendix~\ref{app:seeds}).

\textbf{Models and sampling.}
\ours{} has a 109M block decoder and an 88M prior (197M generation parameters; $d{=}512$, $d_r{=}64$). Its 35M prefix encoder supports training, refinement and prompted continuation (Appendix~\ref{app:prompted}). \ours{}-L uses $d{=}1024$, a 369M decoder, a 308M prior (677M total) and a 109M encoder. Training budgets are 54B and 17B tokens, respectively (ELF-B: 46B; BD3-LM/MDLM: 524B; Appendix~\ref{app:hparams}). Defaults: $K{=}8$ ($\ell{=}128$), 50 prior Euler steps, guidance 1. The \emph{states-only prior} fixes realization codes to zero (Section~\ref{sec:training}); an optional document-wide \emph{4-gram ban} suppresses repetition (Appendix~\ref{app:norepeat}). Table~\ref{tab:main} counts generation parameters, excluding ELF's T5 encoder. \emph{True-plan decode} uses a held-out document's encoded plan.

\textbf{Baselines.}
We evaluate ELF-B/M/L~\citep{hu2026elf} (SDE, guidance 3, 32/64 steps), BD3-LM~\citep{arriola2025block} (blocks of 4/8/16 tokens), and MDLM~\citep{sahoo2024simple} (1000 steps). We include a 197M \emph{same-data AR} (Appendix~\ref{app:ar}) and GPT-2 small~\citep{radford2019language}. Released baselines use official code and checkpoints on our hardware with the same judge; sampling details are in Appendix~\ref{app:baselines}.

\begin{table}[t]
\caption{\textbf{Measured operating points} (settings: Appendix~\ref{app:sweeps}; matched entropy: Figure~\ref{fig:curves}). H100 time: ms/document at batch 32 ($\dagger$: eager; Appendix~\ref{app:batch}).}
\label{tab:main}
\centering
\small
\setlength{\tabcolsep}{2.5pt}
\begin{tabular*}{\linewidth}{@{\extracolsep{\fill}}llrrrrr@{}}
\toprule
Model & Size & TFLOPs & \shortstack{GPU time $\downarrow$\\(ms/doc)} & \genppl{} $\downarrow$ & Entropy & MAUVE $\uparrow$ \\
\midrule
Held-out OpenWebText & --- & --- & --- & 16.9 & 5.41 & --- \\
GPT-2 small (AR) & 124M & 0.27 & 128.6 & 12.09 & 5.122 & 0.75 \\
AR, same data & 197M & 0.43 & 219.2 & 15.37 & 5.151 & 0.95 \\
MDLM & 170M & $\approx$347 & 16\,386.7$^\dagger$ & 35.37 & 4.897 & 0.44 \\
BD3-LM, block 4 & 170M & 17.9 & 1\,416.5$^\dagger$ & 23.75 & 5.039 & 0.61 \\
ELF-B, 32 steps & 105M & 7.2 & 20.9 & 23.97 & 5.158 & 0.73 \\
ELF-B, 64 steps & 105M & 14.2 & 40.9 & 19.42 & 5.096 & 0.89 \\
ELF-M, 64 steps & 342M & 51.2 & 160.2 & 21.65 & 5.184 & 0.88 \\
ELF-L, 64 steps & 652M & 96.4 & 692.3 & 23.65 & 5.281 & 0.84 \\
\midrule
\ours{} $K{=}64$, plain & 197M & 0.80 & 16.0 & 41.15 & 5.183 & 0.87 \\
\ours{} $K{=}32$, plain & 197M & 0.52 & 14.3 & 33.56 & 5.177 & 0.89 \\
\ours{} $K{=}16$, plain & 197M & 0.38 & 16.6 & 25.67 & 5.173 & 0.90 \\
\ours{} $K{=}8$, plain & 197M & 0.31 & 27.3 & 22.75 & 5.164 & 0.93 \\
\ours{} $K{=}8$, states-only & 197M & 0.31 & 32.5 & 18.95 & 5.158 & 0.85 \\
\midrule
\ours{}-L $K{=}16$, plain & 677M & 1.35 & 34.5 & 21.22 & 5.181 & 0.91 \\
\ours{}-L $K{=}8$, plain & 677M & 1.10 & 40.2 & 17.79 & 5.146 & 0.90 \\
\ours{}-L $K{=}8$, states-only & 677M & 1.10 & 45.0 & 15.98 & 5.144 & 0.89 \\
\bottomrule
\end{tabular*}
\vspace{-4pt}
\end{table}

\begin{figure}[t]
\centering
\includegraphics[width=\linewidth]{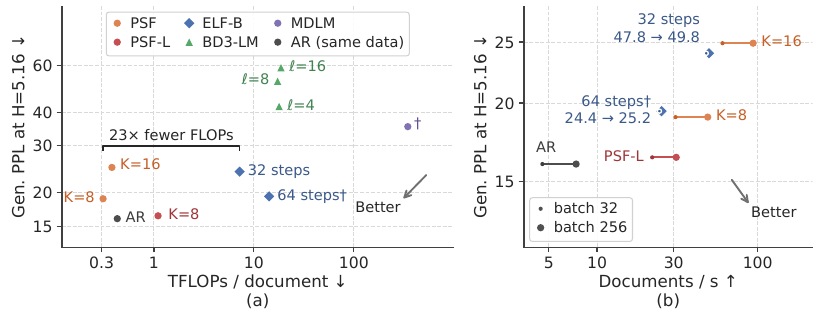}
\vspace{-12pt}
\caption{\textbf{Quality per FLOP and per second} at entropy 5.16; lowest-perplexity setting per \ours{} size and $K$. (b) Lines connect batch 32 (small dots) and 256 (larger markers); ELF labels give documents/s. $\dagger$: own entropy. Sampling and interpolation: Appendix~\ref{app:sweeps}.}
\label{fig:curves}
\vspace{-12pt}
\end{figure}

\subsection{Main results}
\label{sec:results}

\ours{} achieves lower generative perplexity than 32-step ELF-B at matched entropy: at 5.16, the $K{=}8$ states-only configuration with the 4-gram ban reaches 19.0, versus 24.0 for ELF-B (Figure~\ref{fig:curves}(a)). The plain sampler reaches 22.8; states-only sampling without the ban reaches 19.4 (Figure~\ref{fig:landing}(a); Appendix~\ref{app:retrain}). \ours{}-L with the ban reaches 16.4, near the unbanned same-data AR's 16.0 (Appendix~\ref{app:ar}). Perplexity falls as $K$ decreases (Table~\ref{tab:main}; Section~\ref{sec:blocklen}).

Beyond perplexity, the plain $K{=}8$ sampler reaches MAUVE@512 of 0.93 versus 0.73 for 32-step ELF-B, with comparable cross-sample diversity (Appendices~\ref{app:mauve} and \ref{app:memo}). The states-only configuration lowers perplexity further, while the plain sampler retains greater distributional similarity and diversity.

\ours{} retains a 7--23\% perplexity advantage over BD3-LM on a length-controlled window (Appendix~\ref{app:length}). At matched entropy, the best BD3-LM (blocks of 4) is at about twice our perplexity, partly because of its shorter documents. Two Qwen2.5 judges keep the order of the $K{\le}16$ models relative to 32-step ELF-B and BD3-LM (Appendix~\ref{app:judge2}), and a document-wide 4-gram ban reduces the loop rate to zero at similar matched-entropy perplexity (Appendix~\ref{app:norepeat}). Document coherence is evaluated separately (Appendix~\ref{app:coherence}).

\subsection{The landing target}
\label{sec:landing}

\makeatletter\providecommand{\captionof}[1]{\def\@captype{#1}\caption}\makeatother
\definecolor{psfc}{HTML}{DD8452}\definecolor{offc}{HTML}{DA8BC3}\definecolor{nlc}{HTML}{64B5CD}
\begin{figure}[t]
\centering
\includegraphics[width=\linewidth]{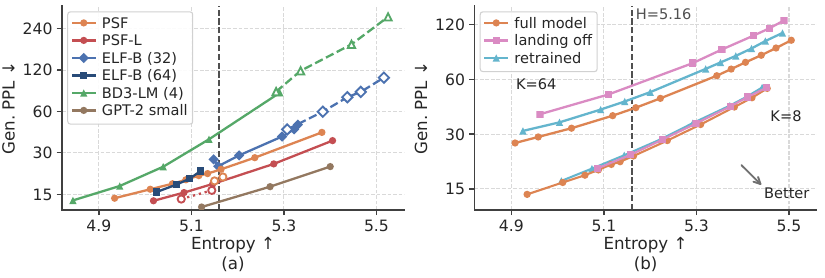}
\vspace{-12pt}
\caption{\textbf{Perplexity--entropy comparisons.} (a) $K{=}8$ plain (solid) and states-only nucleus (hollow, dotted), without the 4-gram ban. ELF-B: 32/64 steps; BD3-LM: block 4. Hollow dashed: baseline guidance/temperature sweeps. (b) Landing-target ablations at $K\in\{8,64\}$; colors as in Table~\ref{tab:seam}.}
\label{fig:landing}
\vspace{0pt}
\begin{minipage}[c]{0.29\linewidth}
\captionof{table}{\textbf{Landing target.} Judge perplexity by position at $K{=}64$ ($\ell{=}16$; geometric means).}
\label{tab:seam}
\end{minipage}\hfill
\begin{minipage}[c]{0.69\linewidth}
\centering
{\footnotesize\setlength{\tabcolsep}{9pt}\renewcommand{\arraystretch}{1.05}
\begin{tabular}{@{}lrrrrr@{}}
\toprule
 & Seam (1) & 2--3 & 4--13 & 14--15 & Last (16) \\
\midrule
Full model \textcolor{psfc}{$\bullet$} & 68.5 & 116 & 30.5 & 44.6 & 84.2 \\
Landing off \textcolor{offc}{$\blacksquare$} & 2293 & 103 & 30.0 & 33.6 & 65.8 \\
Retrained w/o \textcolor{nlc}{$\blacktriangle$} & 379 & 90.0 & 29.6 & 41.6 & 107 \\
\bottomrule
\end{tabular}}
\end{minipage}
\vspace{-12pt}
\end{figure}

The landing target primarily lowers seam perplexity, with larger gains for shorter blocks. It gives each block the next block's first token (Section~\ref{sec:writer}). Without it, the first position of a $K{=}64$ block costs 5.5$\times$ (retrained without the target) to 33$\times$ (switched off at inference) as much as with it, while positions 4--13 stay at about 30 in all three models (Table~\ref{tab:seam}); the target moves some cost to the positions next to the seam. At entropy 5.16, retraining without it costs 14\% with 63 seams (Appendix~\ref{app:landing}) and switching it off costs more, but either costs only 3--6\% with 7 (Figure~\ref{fig:landing}(b)).

\subsection{Efficiency and block length}
\label{sec:compute}

\makeatletter\providecommand{\captionof}[1]{\def\@captype{#1}\caption}\makeatother
\begin{figure}[t]
\begin{minipage}[t]{0.48\linewidth}
\vspace{0pt}
\captionof{table}{\textbf{Predictive vs.\ content states} at $K{=}8$ (\genppl{} \iso; in parentheses: change against predictive). A state that includes its own block is worse in every setting. $^\ddagger$Extrapolated by 0.015 / 0.025 nats (Appendix~\ref{app:content}).}
\label{tab:content}
\vspace{2pt}
\centering\footnotesize\setlength{\tabcolsep}{2.5pt}\renewcommand{\arraystretch}{1.12}
\begin{tabular}{@{}llcr@{\hspace{5.5pt}}l@{}}
\toprule
& & Predictive & \multicolumn{2}{c@{}}{Content} \\
Sampler & Codes & $\vs_k{=}\vc_{k-1}$ & \multicolumn{2}{c@{}}{$\vc_k{=}f_\psi(x^{(\le k)})$} \\
\midrule
plain & prior's & 22.8 & 24.7 & (${+}8\%$) \\
nucleus & prior's & 21.2 & 23.2 & (${+}9\%$) \\
plain & zero & 21.4 & 24.0\rlap{$^\ddagger$} & (${+}12\%$) \\
nucleus & zero & 19.3 & 22.4 & (${+}16\%$) \\
\multicolumn{2}{@{}l}{true-plan decode} & 24.0 & 26.9\rlap{$^\ddagger$} & (${+}12\%$) \\
\midrule
\multicolumn{2}{@{}l}{loop rate ($t{=}0.75$)} & 6.9\% & 7.3\% & \\
\bottomrule
\end{tabular}
\end{minipage}\hfill
\begin{minipage}[t]{0.5\linewidth}
\vspace{0pt}
\centering
\includegraphics[width=\linewidth]{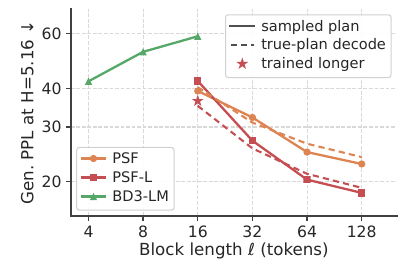}
\vspace{-12pt}
\caption{\textbf{Block length.} Longer blocks help \ours{} but hurt BD3-LM; star: \ours{}-L at $K{=}64$, trained longer (Appendix~\ref{app:blocklen}).}
\label{fig:ablations}
\end{minipage}
\vspace{-12pt}
\end{figure}

\ours{} reduces FLOPs, while its throughput gains depend on batch size and block length. At $K{=}8$, it uses 23$\times$ fewer FLOPs than ELF-B with 32 steps (Figure~\ref{fig:curves}(a)), with 178 serial network evaluations versus ELF-B's 33 and AR's 1024 (Table~\ref{tab:compute}). Figure~\ref{fig:curves}(b) shows \ours{} benefiting more from batching: at batch 256, the selected $K{=}8$ sampler matches ELF-B's throughput at lower perplexity, while $K{=}16$ reaches 1.9$\times$ its throughput. At batch 32, the states-only sampler is 6.7$\times$ faster than the same-data AR, with CUDA graphs on both (Appendix~\ref{app:batch}). BD3-LM and MDLM lack CUDA-graph samplers, so we compare their FLOPs.

\textbf{Block length and scale.}
\label{sec:blocklen}
\label{sec:scaling}
Longer blocks lower generative perplexity and prior cost along the training chain: reducing $K$ from 64 to 8 lowers perplexity from 39.2 to 22.8 and the prior's cost from 631 to 138 GFLOPs, whereas BD3-LM degrades as its blocks grow (Figure~\ref{fig:ablations}). The decoder's held-out cross-entropy rises from 2.63 to 2.97 nats per token (Appendix~\ref{app:blocklen}): predicting a given document and generating text need not improve together. This is consistent with fewer seams and longer within-block context benefiting generation despite coarser states. The larger model helps most with long blocks. We use $K{=}8$, accepting greater serial depth; the chain does not isolate block length at equal training budget.

\textbf{Training and sampling settings.}
\label{sec:knobs}
Training on generated prefixes with source labels brings sampled-plan perplexity close to true-plan decoding (Appendix~\ref{app:landing}). Using 32 or 64 rather than 50 prior steps changes $K{=}8$ headline perplexities by about 1\% (Appendix~\ref{app:nfe}). Results are one-shot:\label{sec:refine} refinement provides no consistent gain at matched entropy (Appendix~\ref{app:refine}).

\subsection{Predictive states and local realization}
\label{sec:ablations}

\textbf{Predictive against content states.}
Predictive states yield lower generative perplexity than content states at matched entropy and $K{=}8$. The content-state control is 8--16\% worse across sampling settings and 12\% worse in true-plan decode, with a similar plain-sampler loop rate (Table~\ref{tab:content}; Appendix~\ref{app:content}). For this control, we replace the predictive state $\vs_k=\vc_{k-1}$ by the content state $\vc_k=f_\psi(x^{(\le k)})$, which also encodes its own block, and repeat the final decoder fine-tuning and prior training at the same budget. The gap persists when the prior is bypassed, suggesting that the benefit depends partly on how the decoder uses the representation. Both variants inherit a predictive-state checkpoint, so the result supports this division of work under the shared adaptation protocol.

\textbf{Realization codes and the states-only prior.}
At matched entropy, generated realization codes lower generative perplexity at the tested $K{\ge}16$ settings but raise it at $K{=}8$. With code-trained decoders, zeroing the codes raises perplexity by 5--38\% at $K{\ge}16$, but lowers it by 6\% (plain) and 9\% (nucleus) at $K{=}8$ (Appendix~\ref{app:infabl}). A states-only prior retains the latter gains (Appendix~\ref{app:retrain}). Longer blocks provide more local history, consistent with leaving more realization to the decoder. We therefore generate codes at $K{\ge}16$ and use states-only generation at $K{=}8$.

\textbf{What the sampled plan controls.}
\label{sec:plancontrol}
Sampled plans yield higher MAUVE@512 than a shared mean plan at $K{=}8$: 0.866 versus 0.754 at a fixed sampler setting (Appendix~\ref{app:planprobe}). Replacing sampled plans by their position-wise mean, with zero codes and fixed decoder noise, also raises \ours{}'s perplexity by 7--9\% and the lexical topic-switch rate from 12--14\% to 30--47\%. Perplexity is compared at fixed sampler settings; lexical topic-switch rates are interpolated to entropy 5.16. The intervention also changes length and entropy. Re-encoded outputs are closer to their own sampled plan than to another document's in 98\% of cases. These probes support the plan's role in organizing content; similar sampled-plan and true-plan perplexities alone would not establish that role.

\textbf{Quality of the best configuration.}
\label{sec:quality}
Our lowest-perplexity configuration in Table~\ref{tab:main}, \ours{}-L with the states-only prior, still has a higher lexical topic-switch rate than real text: 9.6\% versus 3.6\%. Appendices~\ref{app:topic} and \ref{app:coherence} report topic-persistence and LLM-judged coherence evaluations.

\section{Background and Related Work}

\textbf{Diffusion and flow models for text.}
Discrete DLMs corrupt tokens with multinomial or masking noise and learn to reverse it~\citep{austin2021structured,lou2023discrete,sahoo2024simple}; continuous DLMs embed tokens and denoise the embeddings~\citep{li2022diffusion,gulrajani2023likelihood}. ELF~\citep{hu2026elf} keeps the sequence in a frozen T5 embedding space and reaches a generative perplexity of 24 on OpenWebText in 32 steps; our prior borrows its self-conditioning~\citep{chen2022analog} and guidance-scale tokens~\citep{ho2022classifier}. All of these models denoise the full $L$-token sequence at every step; our prior denoises $K \ll L$ latents, and tokens are produced by a separate block decoder.

\textbf{Block-wise and semi-autoregressive generation.}
BD3-LM~\citep{arriola2025block}, SSD-LM~\citep{han2023ssd} and AR-Diffusion~\citep{wu2023ar} are autoregressive \emph{across} blocks and diffuse \emph{within} each block; BD3-LM's quality degrades as the block grows (Section~\ref{sec:blocklen}). Our model is the transpose---diffusion across blocks, autoregression within---so the two families live on one axis, the block length, and we report both ends of it under one ruler. Blockwise parallel, multi-head and speculative decoding~\citep{stern2018blockwise,cai2024medusa,leviathan2023fast} propose tokens and verify them with an autoregressive model; our blocks need no verification step: there is no separate target model to match, and the plan and the landing token are the only cross-block inputs.

\textbf{Planning and hierarchical generation.}
Plan-and-Write~\citep{yao2019plan} uses explicit keyword storylines to guide story generation, while Skeleton-of-Thought~\citep{ning2024skeleton} expands a prompted textual outline in parallel. The Latent Transformer~\citep{kaiser2018fast} autoregressively generates a shorter sequence of discrete latents and decodes the target translation in parallel. Our flow prior generates continuous predictive states jointly, and the block decoder realizes tokens autoregressively within each block.

\textbf{Latent diffusion for text.}
LD4LG~\citep{lovelace2023latent} and PLANNER~\citep{zhang2023planner} run diffusion in a learned latent space of a sentence or paragraph and decode it autoregressively; Large Concept Models~\citep{barrault2024large} generate sentence embeddings and decode each into its sentence. These representations encode the content to be generated. We generate a \emph{sequence} of predictive states that exclude their own block, predict the blocks \emph{in parallel}, and couple adjacent blocks with a one-token landing target.

\section{Conclusion}




This work revisits what a diffusion prior should generate in a continuous diffusion language model. 
Rather than requiring continuous states to specify the final token sequence, \method{} assigns different generative roles to the prior and the decoder: the flow prior generates a short trajectory of predictive states that describe what should come next, while a block decoder resolves the corresponding token-level realizations. 
One-token landing targets provide lightweight coupling between blocks while preserving parallel block generation.
This division of roles yields both computational and modeling benefits. With only $K=8$ predictive states for a 1024-token document, \ours{} uses 23$\times$ fewer FLOPs than 32-step ELF-B while achieving lower generative perplexity, higher MAUVE, and higher batched throughput. 
When the prior generates predictive states alone, nucleus sampling further reduces matched-entropy perplexity to approximately 20\% below ELF-B. 
More importantly, predictive states consistently outperform content states that additionally encode their own blocks, and increasing block length improves \ours{} even as it degrades block diffusion.

Together, these results suggest that continuous language-model latents should be designed not only for what they represent, but also for how generative roles are divided between the prior and the decoder.
A useful latent need not preserve every detail of the output; it can instead represent information that is easy for a continuous prior to generate and sufficient for a decoder to turn into coherent text. Predictive states provide one realization of this principle, and point toward continuous generative models in which priors and decoders are designed jointly around the information each is best suited to generate.

\textbf{Limitations and future work.}\label{sec:limitations}
Our main limitation is scale: all experiments use at most 677M generation parameters, a fixed 1024-token canvas and unconditional OpenWebText generation; larger models and longer sequences remain unexplored. Blocks are decoded in parallel from a shared plan, coupled only by one-token landing targets, so whether many blocks stay consistent in topic, entities and discourse over longer documents is untested; topic persistence is already the weakest aspect of our samples (Appendix~\ref{app:topic}). The most pressing next steps are to scale model size and sequence length, to couple blocks more strongly than through a single token, and to measure long-range coherence, which generative perplexity misses. Prompted continuation works without retraining but uses the prompt weakly, calling for a prompt-conditioned prior. Autoregressive models remain ahead, although the gap to the same-data baseline narrows at our largest scale (Appendix~\ref{app:ar}), and the states-only configuration trades topic persistence, diversity and judged coherence for lower perplexity. The FLOP saving over ELF becomes a throughput advantage only at large batch, and the training recipe has several stages, including realization codes whose benefit we only partly understand; faster small-batch sampling and a simpler recipe remain future work.

\bibliography{dlm}

@article{hu2026elf,
  title={Elf: Embedded language flows},
  author={Hu, Keya and Qiu, Linlu and Lu, Yiyang and Zhao, Hanhong and Li, Tianhong and Kim, Yoon and Andreas, Jacob and He, Kaiming},
  journal={arXiv preprint arXiv:2605.10938},
  year={2026}
}

@inproceedings{arriola2025block,
  title={Block diffusion: Interpolating between autoregressive and diffusion language models},
  author={Arriola, Marianne and Gokaslan, Aaron and Chiu, Justin and Yang, Zhihan and Qi, Zhixuan and Han, Jiaqi and Sahoo, Subham and Kuleshov, Volodymyr},
  booktitle={International Conference on Learning Representations},
  volume={2025},
  pages={50726--50753},
  year={2025}
}

@article{sahoo2024simple,
  title={Simple and effective masked diffusion language models},
  author={Sahoo, Subham S and Arriola, Marianne and Schiff, Yair and Gokaslan, Aaron and Marroquin, Edgar and Chiu, Justin T and Rush, Alexander and Kuleshov, Volodymyr},
  journal={Advances in Neural Information Processing Systems},
  volume={37},
  pages={130136--130184},
  year={2024}
}

@article{lou2023discrete,
  title={Discrete diffusion modeling by estimating the ratios of the data distribution},
  author={Lou, Aaron and Meng, Chenlin and Ermon, Stefano},
  journal={arXiv preprint arXiv:2310.16834},
  year={2023}
}

@article{austin2021structured,
  title={Structured denoising diffusion models in discrete state-spaces},
  author={Austin, Jacob and Johnson, Daniel D and Ho, Jonathan and Tarlow, Daniel and Van Den Berg, Rianne},
  journal={Advances in neural information processing systems},
  volume={34},
  pages={17981--17993},
  year={2021}
}

@article{li2022diffusion,
  title={Diffusion-lm improves controllable text generation},
  author={Li, Xiang and Thickstun, John and Gulrajani, Ishaan and Liang, Percy S and Hashimoto, Tatsunori B},
  journal={Advances in neural information processing systems},
  volume={35},
  pages={4328--4343},
  year={2022}
}

@article{gulrajani2023likelihood,
  title={Likelihood-based diffusion language models},
  author={Gulrajani, Ishaan and Hashimoto, Tatsunori B},
  journal={Advances in Neural Information Processing Systems},
  volume={36},
  pages={16693--16715},
  year={2023}
}

@article{lovelace2023latent,
  title={Latent diffusion for language generation},
  author={Lovelace, Justin and Kishore, Varsha and Wan, Chao and Shekhtman, Eliot and Weinberger, Kilian Q},
  journal={Advances in Neural Information Processing Systems},
  volume={36},
  pages={56998--57025},
  year={2023}
}

@article{zhang2023planner,
  title={Planner: Generating diversified paragraph via latent language diffusion model},
  author={Zhang, Yizhe and Gu, Jiatao and Wu, Zhuofeng and Zhai, Shuangfei and Susskind, Joshua and Jaitly, Navdeep},
  journal={Advances in Neural Information Processing Systems},
  volume={36},
  pages={80178--80190},
  year={2023}
}

@article{liu2022flow,
  title={Flow straight and fast: Learning to generate and transfer data with rectified flow},
  author={Liu, Xingchao and Gong, Chengyue and Liu, Qiang},
  journal={arXiv preprint arXiv:2209.03003},
  year={2022}
}

@article{lipman2022flow,
  title={Flow matching for generative modeling},
  author={Lipman, Yaron and Chen, Ricky TQ and Ben-Hamu, Heli and Nickel, Maximilian and Le, Matt},
  journal={arXiv preprint arXiv:2210.02747},
  year={2022}
}

@article{ho2022classifier,
  title={Classifier-free diffusion guidance},
  author={Ho, Jonathan and Salimans, Tim},
  journal={arXiv preprint arXiv:2207.12598},
  year={2022}
}

@article{chen2022analog,
  title={Analog bits: Generating discrete data using diffusion models with self-conditioning},
  author={Chen, Ting and Zhang, Ruixiang and Hinton, Geoffrey},
  journal={arXiv preprint arXiv:2208.04202},
  year={2022}
}

@article{radford2019language,
  title={Language models are unsupervised multitask learners},
  author={Radford, Alec and Wu, Jeffrey and Child, Rewon and Luan, David and Amodei, Dario and Sutskever, Ilya and others},
  journal={OpenAI blog},
  volume={1},
  number={8},
  pages={9},
  year={2019}
}

@misc{Gokaslan2019OpenWeb,  
	title={OpenWebText Corpus},
	author={Aaron Gokaslan and Vanya Cohen},
	howpublished={\url{http://Skylion007.github.io/OpenWebTextCorpus}}, 
	year={2019}
}

@article{raffel2020exploring,
  title={Exploring the limits of transfer learning with a unified text-to-text transformer},
  author={Raffel, Colin and Shazeer, Noam and Roberts, Adam and Lee, Katherine and Narang, Sharan and Matena, Michael and Zhou, Yanqi and Li, Wei and Liu, Peter J},
  journal={Journal of machine learning research},
  volume={21},
  number={140},
  pages={1--67},
  year={2020}
}

@article{holtzman2019curious,
  title={The curious case of neural text degeneration},
  author={Holtzman, Ari and Buys, Jan and Du, Li and Forbes, Maxwell and Choi, Yejin},
  journal={arXiv preprint arXiv:1904.09751},
  year={2019}
}

@inproceedings{han2023ssd,
  title={Ssd-lm: Semi-autoregressive simplex-based diffusion language model for text generation and modular control},
  author={Han, Xiaochuang and Kumar, Sachin and Tsvetkov, Yulia},
  booktitle={Proceedings of the 61st Annual Meeting of the Association for Computational Linguistics (Volume 1: Long Papers)},
  pages={11575--11596},
  year={2023}
}

@article{wu2023ar,
  title={Ar-diffusion: Auto-regressive diffusion model for text generation},
  author={Wu, Tong and Fan, Zhihao and Liu, Xiao and Zheng, Hai-Tao and Gong, Yeyun and Jiao, Jian and Li, Juntao and Guo, Jian and Duan, Nan and Chen, Weizhu and others},
  journal={Advances in Neural Information Processing Systems},
  volume={36},
  pages={39957--39974},
  year={2023}
}

@article{stern2018blockwise,
  title={Blockwise parallel decoding for deep autoregressive models},
  author={Stern, Mitchell and Shazeer, Noam and Uszkoreit, Jakob},
  journal={Advances in Neural Information Processing Systems},
  volume={31},
  year={2018}
}

@inproceedings{leviathan2023fast,
  title={Fast inference from transformers via speculative decoding},
  author={Leviathan, Yaniv and Kalman, Matan and Matias, Yossi},
  booktitle={International conference on machine learning},
  pages={19274--19286},
  year={2023},
  organization={PMLR}
}

@article{pillutla2021mauve,
  title={Mauve: Measuring the gap between neural text and human text using divergence frontiers},
  author={Pillutla, Krishna and Swayamdipta, Swabha and Zellers, Rowan and Thickstun, John and Welleck, Sean and Choi, Yejin and Harchaoui, Zaid},
  journal={Advances in Neural Information Processing Systems},
  volume={34},
  pages={4816--4828},
  year={2021}
}

@article{meshchaninov2026cosmos,
  title={Cosmos: Compressed and smooth latent space for text diffusion modeling},
  author={Meshchaninov, Viacheslav and Chimbulatov, Egor and Shabalin, Alexander and Abramov, Aleksandr and Vetrov, Dmitry},
  journal={Advances in Neural Information Processing Systems},
  volume={38},
  pages={14271--14299},
  year={2026}
}

@article{cai2024medusa,
  title={Medusa: Simple llm inference acceleration framework with multiple decoding heads},
  author={Cai, Tianle and Li, Yuhong and Geng, Zhengyang and Peng, Hongwu and Lee, Jason D and Chen, Deming and Dao, Tri},
  journal={arXiv preprint arXiv:2401.10774},
  year={2024}
}

@inproceedings{ning2024skeleton,
  title={Skeleton-of-thought: Prompting llms for efficient parallel generation},
  author={Ning, Xuefei and Lin, Zinan and Zhou, Zixuan and Wang, Zifu and Yang, Huazhong and Wang, Yu},
  booktitle={International Conference on Learning Representations},
  volume={2024},
  pages={917--967},
  year={2024}
}

@article{barrault2024large,
  title={Large concept models: Language modeling in a sentence representation space},
  author={Barrault, Lo{\"\i}c and Duquenne, Paul-Ambroise and Elbayad, Maha and Kozhevnikov, Artyom and Alastruey, Belen and Andrews, Pierre and Coria, Mariano and Couairon, Guillaume and Costa-juss{\`a}, Marta R and Dale, David and others},
  journal={arXiv preprint arXiv:2412.08821},
  year={2024}
}

@inproceedings{yao2019plan,
  title={Plan-and-write: Towards better automatic storytelling},
  author={Yao, Lili and Peng, Nanyun and Weischedel, Ralph and Knight, Kevin and Zhao, Dongyan and Yan, Rui},
  booktitle={Proceedings of the AAAI conference on artificial intelligence},
  volume={33},
  number={01},
  pages={7378--7385},
  year={2019}
}

@inproceedings{kaiser2018fast,
  title={Fast decoding in sequence models using discrete latent variables},
  author={Kaiser, Lukasz and Bengio, Samy and Roy, Aurko and Vaswani, Ashish and Parmar, Niki and Uszkoreit, Jakob and Shazeer, Noam},
  booktitle={International Conference on Machine Learning},
  pages={2390--2399},
  year={2018},
  organization={PMLR}
}

@inproceedings{rombach2022high,
  title={High-resolution image synthesis with latent diffusion models},
  author={Rombach, Robin and Blattmann, Andreas and Lorenz, Dominik and Esser, Patrick and Ommer, Bj{\"o}rn},
  booktitle={2022 IEEE/CVF conference on computer vision and pattern recognition (CVPR)},
  pages={10674--10685},
  year={2022},
  organization={ieee}
}

@article{nie2026large,
  title={Large language diffusion models},
  author={Nie, Shen and Zhu, Fengqi and You, Zebin and Zhang, Xiaolu and Ou, Jingyang and Hu, Jun and Zhou, Jun and Lin, Yankai and Wen, Ji-Rong and Li, Chongxuan},
  journal={Advances in Neural Information Processing Systems},
  volume={38},
  pages={50608--50646},
  year={2026}
}

@article{dieleman2022continuous,
  title={Continuous diffusion for categorical data},
  author={Dieleman, Sander and Sartran, Laurent and Roshannai, Arman and Savinov, Nikolay and Ganin, Yaroslav and Richemond, Pierre H and Doucet, Arnaud and Strudel, Robin and Dyer, Chris and Durkan, Conor and others},
  journal={arXiv preprint arXiv:2211.15089},
  year={2022}
}

@inproceedings{hu2024flow,
  title={Flow matching for conditional text generation in a few sampling steps},
  author={Hu, Vincent and Wu, Di and Asano, Yuki and Mettes, Pascal and Fernando, Basura and Ommer, Bj{\"o}rn and Snoek, Cees},
  booktitle={Proceedings of the 18th Conference of the European Chapter of the Association for Computational Linguistics (Volume 2: Short Papers)},
  pages={380--392},
  year={2024}
}

@inproceedings{esser2024scaling,
  title={Scaling rectified flow transformers for high-resolution image synthesis},
  author={Esser, Patrick and Kulal, Sumith and Blattmann, Andreas and Entezari, Rahim and M{\"u}ller, Jonas and Saini, Harry and Levi, Yam and Lorenz, Dominik and Sauer, Axel and Boesel, Frederic and others},
  booktitle={Forty-first international conference on machine learning},
  year={2024}
}
\bibliographystyle{iclr2027_conference}

\makeatletter\setlength{\@fptop}{0pt}\setlength{\@fpsep}{8pt}\setlength{\@fpbot}{0pt plus 1fil}\makeatother\renewcommand{\floatpagefraction}{0.85}  

\clearpage

\appendix

\startcontents[appendix]
\section*{Appendix}

\noindent\textbf{Terminology.} When describing \method{}, we use \emph{diffusion} and \emph{flow} interchangeably to refer to the same rectified-flow prior and its ODE sampling process.

\printcontents[appendix]{}{1}[2]{}

\newpage
\section{Experimental details}
\label{app:details}

This appendix records the architecture, training schedules, evaluation protocol and sampling points needed to reproduce the results.

\subsection{Architecture and information flow}
\label{app:architecture}

\begin{figure}[!htbp]
\centering
\begin{minipage}{0.55\linewidth}
\input{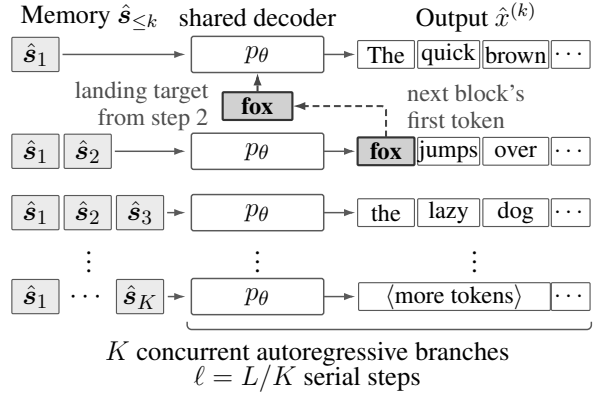}%
\end{minipage}
\caption{\textbf{Parallel block decoding.} Branch $k$ reads the cumulative memory $\hat{\vs}_{\le k}$ and generates its block autoregressively. All branches share parameters and advance synchronously for $\ell=L/K$ steps. The dashed path shows a landing target: after the first step, the sampled first token of the next block (``fox'' here) conditions the preceding branch from step 2 onward.}
\label{fig:block-decoder}
\end{figure}

Figures~\ref{fig:prefix-encoder}--\ref{fig:block-decoder} trace information flow through \ours{}: each predictive state is constructed from its preceding blocks, the prior jointly generates all states, and each decoder branch reads its cumulative state memory plus the next block's first token after the initial decoding step. The diagrams show the states-only path; Appendix~\ref{app:codes} describes the optional realization-code inputs.

\begin{figure}[!htbp]
\centering
\begin{minipage}[t]{0.48\linewidth}
\centering
\vspace{0pt}
\input{fig_prefix_encoder}%
\caption{\textbf{Predictive-state construction.} A single block-causal encoder pass produces pooled states $\vc_j$. Shifting by one block gives $\vs_1=\vs_{\mathrm{bos}}$ and $\vs_k=\vc_{k-1}$ for $k\ge2$. The prefix strips show exactly which blocks each state can observe.}
\label{fig:prefix-encoder}
\end{minipage}\hfill
\begin{minipage}[t]{0.48\linewidth}
\centering
\vspace{0pt}
\begin{minipage}{0.81\linewidth}
\input{fig_flow_prior}%
\end{minipage}
\caption{\textbf{Joint state generation.} The flow prior transforms Gaussian noise into the full trajectory $\hat{\mathbf S}=(\hat{\vs}_1,\dots,\hat{\vs}_K)$ in $T$ sampling steps. Bidirectional attention couples all $K$ states; the prior operates on this short state sequence.}
\label{fig:flow-prior}
\end{minipage}
\end{figure}

\FloatBarrier

\subsection{Architectures, training schedule and optimizer}
\label{app:hparams}

Table~\ref{tab:arch} specifies the three networks. Table~\ref{tab:stages} then gives the model-specific training schedules, followed by the shared optimizer and prior settings in Table~\ref{tab:optim}.

\begin{table}[!htbp]
\caption{Architecture of the three networks. Widths and parameter counts are given as \ours{} / \ours{}-L ($d = 512$ / $1024$).}
\label{tab:arch}
\centering
\footnotesize\setlength{\tabcolsep}{3pt}
\begin{tabular}{l>{\raggedright\arraybackslash}p{0.9cm}>{\raggedright\arraybackslash}p{1.3cm}>{\raggedright\arraybackslash}p{0.8cm}>{\raggedright\arraybackslash}p{1.5cm}>{\raggedright\arraybackslash}p{2.7cm}>{\raggedright\arraybackslash}p{3.0cm}}
\toprule
Network & Layers & Width & Heads & Params & Attention and positions & Inputs and outputs \\
\midrule
Encoder $E_\psi$ & 6 & 512 / 1024 & 8 & 35M / 109M & block-causal self-attention; rotary embeddings with a hybrid within-block / across-block scheme & per-block mean pooling; linear head to the code posterior $(\vmu_k, \log\vsigma_k)$, $d_r = 64$ \\
Block decoder $p_\theta$ & 18 & 512 / 1024 & 8 & 109M / 369M & causal self-attention within a block; cross-attention to $\vs_{\le k}$ with a causal mask over blocks; rotary embeddings within blocks & added at every position: $\vs_k$-conditioned memory, $\vr_k$ and $\vr_{k+1}$ (zero-initialized linear layers), the source label $\lambda_k$ (zero-initialized 2-dimensional embedding), and from position 2 the landing embedding of $x^{(k+1)}_1$ \\
Prior $p_\phi$ & 12 / 24 & 768 / 1024 & 12 / 16 & 88M / 308M & full self-attention over $K$ tokens & tokens of dimension $d + d_r$ through a 256-dimensional input bottleneck; 4 guidance-scale tokens and 4 time tokens \\
\bottomrule
\end{tabular}

\end{table}

\begin{table}[!htbp]
\caption{Training schedule of the encoder and block decoder (Stages I--II of Section~\ref{sec:training}) and the prior (Stage III). Batch = documents per step (1024-token canvas); KL budget = code budget in nats per block ($\ell$ nats, i.e.\ one nat per token; ``---'' = no code channel yet); tokens = steps $\times$ batch $\times$ 953 non-padding tokens. Notes below the table say what each stage adds and when the encoder is frozen.}
\label{tab:stages}
\centering
\small
\begin{tabular}{lrrrrr}
\toprule
Stage & Steps (k) & Batch & $K$ & KL budget & Tokens (B) \\
\midrule
\multicolumn{6}{l}{\emph{\ours{}, Stages I--II}} \\
base (Stage I)$^{a}$ & 30 & 512 & 64 & --- & 14.6 \\
landing target$^{b}$ & 10 & 512 & 64 & --- & 4.9 \\
full recipe$^{c}$ & 30 & 512 & 64 & --- & 14.6 \\
self-realized prefixes$^{d}$ & 10 & 512 & 64 & --- & 4.9 \\
realization codes$^{e}$ & 10 & 512 & 64 & 16 & 4.9 \\
chain $K{=}32$$^{f}$ & 8 & 512 & 32 & 32 & 3.9 \\
chain $K{=}16$$^{f}$ & 6 & 512 & 16 & 64 & 2.9 \\
chain $K{=}8$$^{f}$ & 6 & 512 & 8 & 128 & 2.9 \\
total & 110 & & & & 54 \\
\addlinespace[3pt]
\multicolumn{6}{l}{\emph{\ours{}-L ($d{=}1024$), Stages I--II}} \\
base (Stage I)$^{a}$ & 39.5 & 256 & 64 & --- & 9.6 \\
self-realized prefixes$^{d}$ & 12 & 192 & 64 & --- & 2.2 \\
realization codes$^{e}$ & 10 & 192 & 64 & 16 & 1.8 \\
chain $K{=}32$$^{f}$ & 8 & 192 & 32 & 32 & 1.5 \\
chain $K{=}16$$^{f}$ & 6 & 192 & 16 & 64 & 1.1 \\
chain $K{=}8$$^{f}$ & 6 & 192 & 8 & 128 & 1.1 \\
total & 81.5 & & & & 17 \\
\addlinespace[3pt]
\multicolumn{6}{l}{\emph{Prior (Stage III), one run per block length}$^{g}$} \\
\ours{}, each $K$ & 30 & 512 & 64 / 32 / 16 / 8 & & \\
\ours{}-L, each $K$ & 30 & 384 & 64 / 32 / 16 / 8 & & \\
\bottomrule
\multicolumn{6}{@{}p{0.97\textwidth}@{}}{\footnotesize
$^{a}$~Encoder and block decoder from scratch with token cross-entropy; the only stage that trains the encoder trunk. \ours{}-L uses landing targets from the start, without self-realized prefixes. Its encoder stays bit-identical through stage~$d$ because the states are detached.
$^{b}$~For \ours{}, adds the landing target (Sec.~\ref{sec:writer}) after Stage I; the encoder stays fixed.
$^{c}$~The remaining Sec.~\ref{sec:method} recipe except self-realized prefixes and codes.
$^{d}$~Self-realized prefix training with the source label $\lambda_k$ (Sec.~\ref{sec:training}).
$^{e}$~Realization codes $\vr_k$ with lookahead $\vr_{k+1}$ (Sec.~\ref{sec:training}); the code head, a linear read-out, trains on the frozen trunk from here on.
$^{f}$~Block structure and KL budget only; encoder frozen.
$^{g}$~Warm-started from the previous block length's prior; the $K{=}64$ prior of \ours{} ran 60k steps.} \\
\end{tabular}

\caption{Optimizer, regularizers and the prior objective. Values are shared by \ours{} and \ours{}-L unless stated.}
\label{tab:optim}
\centering
\small
\begin{tabular}{l>{\raggedright\arraybackslash}p{7.8cm}}
\toprule
Optimizer (encoder, block decoder, prior) & AdamW, $\beta = (0.9, 0.95)$, weight decay 0.01 (none on norms and biases) \\
Learning rate & peak $3 \cdot 10^{-4}$, 1000 warm-up steps, cosine decay to 10\% \\
Gradient clipping & 1.0 \\
Precision & bfloat16 autocast \\
EMA of weights & block decoder 0.999, prior 0.9999 \\
Code regularizer & Kullback--Leibler to $\mathcal{N}(\mathbf{0}, \mI)$, adaptive coefficient, target $\ell$ nats per block \\
Block-decoder noise & Gaussian noise added to the predictive states the block decoder reads during training \\
Self-realized prefixes & states replaced per block with probability 0.5; rollout at temperature 0.65, nucleus 0.95; original posterior codes retained \\
Prior objective & rectified flow; $x$-prediction with a velocity loss, $\vz_t = t\,\vz_1 + (1-t)\,\vepsilon$ \\
Prior time distribution & logit-normal, mean $-1.5$, standard deviation $0.8$ \\
Prior self-conditioning & probability 0.5 \\
Prior guidance scale & drawn log-uniformly in $[0.5, 5]$ during training; 1 at sampling unless stated \\
Prior code target & posterior means $\vmu_k$ (states-only prior: zero) \\
Prior sampler & 50 Euler steps of the probability-flow ODE; prior noise scale = standard deviation of the initial $\vepsilon$, 1 by default \\
\bottomrule
\end{tabular}

\end{table}

\paragraph{Schedule differences, freezing and warm starts.}
\ours{} introduces landing targets after base training, whereas \ours{}-L uses them from the start and proceeds directly to self-realized prefixes. Both models freeze the encoder trunk after the base stage; later encoder-side updates are confined to the added code head. Thus their $K = 32, 16, 8$ predictive states all use the encoder trained at $K = 64$, including the retrained ablations of Appendix~\ref{app:retrain} and continued training of Appendix~\ref{app:bigcont}. Each prior is warm-started from the previous block length's prior; tensors whose shape depends on the latent dimension are slice-copied when $d$ changes.

\begin{samepage}
\paragraph{Auxiliary realization-code extension.}
\label{app:codes}
\emph{Posterior training.} A second encoder head defines $q_\psi(\vr_k\mid x^{(\le k)})=\mathcal{N}(\vmu_k,\mathrm{diag}\,\vsigma_k^2)$, with $\vr_k\in\R^{d_r}$ and $d_r=64$. Block-decoder training samples from this posterior and regularizes it toward $\mathcal{N}(\mathbf{0},\mI)$ with an adaptive Kullback--Leibler budget of $\ell$ nats per block (one nat per token).

\emph{Decoder inputs.} Zero-initialized linear layers add $\vr_k$ and the next block's code $\vr_{k+1}$ at every position, with $\vr_{K+1}=\mathbf{0}$. These inputs augment \eqref{eq:writer}; the landing mechanism is unchanged.

\emph{Prior targets.} The code-augmented prior fits posterior means with clean plan tokens $\vz_k=[\vs_k;\vmu_k]$. The states-only prior uses $\vz_k=[\vs_k;\mathbf{0}]$, retaining $d+d_r$ coordinates (576 in \ours{}) with zero clean code targets. It retains the code-trained decoder and explicitly zeros its code inputs at generation.
\par\end{samepage}

\paragraph{Compute and software.}
All training ran on eight H100 80GB GPUs in PyTorch with bfloat16 autocast. Step cost is dominated by the self-realized-prefix rollouts. The timed stages are: \ours{}'s chain stage at $K{=}8$, 6.6 s per step (11 hours for 6k steps); for \ours{}-L, the realization-code stage 1.1 s per step (3.1 hours for 10k steps), the chain stages at $K{=}32$, $16$ and $8$ 1.7, 3.0 and 5.6 s per step (3.9, 5.0 and 9.4 hours for 8k, 6k and 6k steps), and a 30k-step prior stage 0.21 s per step (1.8 hours). The remaining stages of Table~\ref{tab:stages} were not timed. Every inference timing uses one otherwise idle H100 (Appendix~\ref{app:batch}).

\FloatBarrier
\subsection{Evaluation protocol}
\label{app:protocol}
\label{app:metrics}

The following protocol defines the quality columns of Table~\ref{tab:main} and the interpolated comparisons.

\emph{Text processing and held-out data.} We detokenize each generated document with its native tokenizer (T5, \texttt{google-t5/t5-small}, $V = 32{,}100$ for \ours{} and ELF), cut at the first end-of-text token, and re-tokenize with \texttt{openai-community/gpt2-large}. Scoring uses at most the first 1024 judge tokens per document. A fixed random subset of 10{,}000 OpenWebText documents (split seed 0) is held out from training. Evaluation dumps and the appendix's held-out rows each contain 1000 documents.

\emph{Perplexity, entropy and repetition.} Generative perplexity is the exponential of the corpus-level mean token negative log-likelihood. Entropy is the unigram entropy of the judge's tokens, computed per document and averaged in nats; on ELF's samples our implementation and ELF's agree to the third decimal. Loop rate is the fraction of documents whose first 256 T5 tokens contain an 8-gram occurring at least three times.

\emph{MAUVE.} MAUVE~\citep{pillutla2021mauve} compares 1000 generated and 1000 held-out documents using GPT-2 Large's final-layer state at the last token, with $k$-means quantization and five quantization seeds averaged. This feature is sensitive to whether a document ends naturally or is cut at the canvas. Systems that never emit end-of-text (ELF and GPT-2 small at a fixed 1024 tokens) score 0.02--0.46 with the full 1024-token window, reflecting a length artefact (Appendix~\ref{app:length}). We therefore restrict both sides to their first 512 judge tokens by default. Cosmos uses 256 on both sides (Appendix~\ref{app:cosmos}); scores from different windows are not directly comparable.

\emph{Interpolation.} We obtain ``\iso'' values by log-linear interpolation to entropy 5.16, the rounded entropy of ELF-B's 32-step samples (5.158), with any extrapolations explicitly marked. Perplexity--entropy slopes are 2.2--2.5 log-perplexity units per nat for our models: about 25\% higher perplexity per 0.1 nat ($e^{0.22}\approx1.25$). The slopes are about 3.4 for ELF-B (Table~\ref{tab:elfsteps}) and 5--9 for BD3-LM (Table~\ref{tab:bd3sweep}), making the choice of bracket especially consequential for BD3-LM. Appendix~\ref{app:sweeps} lists the brackets and distinguishes measured from interpolated points.

\emph{Other judges and diagnostics.} Appendix~\ref{app:judge2} uses \texttt{Qwen/Qwen2.5-1.5B} and \texttt{Qwen/Qwen2.5-7B} under the same protocol, with entropy in Qwen tokens. True-plan decode replaces the sampled plan by the encoder's plan of a held-out document, assessing the decoder with the prior bypassed. Size in Table~\ref{tab:main} counts generation parameters (decoder plus prior; ELF excludes T5).

\paragraph{Likelihood bound.}
\label{app:elbo}
The likelihood bound provides a density-based diagnostic, but is loose and does not replace our generative-perplexity comparison, as in ELF. \ours{} has a continuous latent prior: we evaluate an ELBO using a Gaussian posterior around the encoder's states and code means, stopping the prior ODE at $t^{\ast}<1$. The decoder reads a rescaled state with residual noise, consistent with its noise-augmented training. The prior term is the exact probability-flow ODE density at $t^{\ast}$; evaluation uses 1024 held-out documents and nats per real T5 token. For $K{=}8$ at $t^{\ast}{=}0.7$, the decoder term is 3.100, the prior term 4.013 and $-H(q)$ is $-2.745$, giving $-$ELBO $=4.37$ (5.20 at $t^{\ast}{=}0.8$). At $t^{\ast}{=}0.7$, $K{=}16$ gives 5.60 and $K{=}64$ gives 11.8, compared with held-out cross-entropy 2.635 for the same-data AR (Appendix~\ref{app:ar}). The bound includes a prior density over $K(d+d_r)=4608$ latent coordinates per 953-token document. The ordering across $K$ agrees with generative perplexity. The decoder's teacher-forced negative log-likelihood (Table~\ref{tab:blocklen}) remains a component diagnostic.

\subsection{Baselines}
\label{app:baselines}

We sample the baselines on our hardware using their official code and released checkpoints, retaining each model's native sampler and precision (Table~\ref{tab:baselines}). We then detokenize and score them with the common judge of Appendix~\ref{app:metrics}.

Published values need not use this text-processing protocol. BD3-LM and MDLM fill a 1024-token canvas that may contain an end-of-text token followed by another document. Their papers score the whole canvas; we score its first document, as in ELF's evaluation, lowering their entropy by about 0.3 nats and their perplexity accordingly. Our entropy is averaged per document, matching ELF's definition. Table~\ref{tab:baselines} places published and re-scored results side by side; comparisons use the latter. Additional ELF-B step and noise settings appear in Appendix~\ref{app:elfsteps}, and Cosmos's different native length is documented below.

\begin{table}[!htbp]
\caption{Baselines: sampling settings, published generative perplexity and entropy, and the same quantities under our protocol (1000 documents, GPT-2 Large judge, first document of each canvas). The BD3-LM rows at nucleus 0.95 and 1.0 complete the sweep behind the log-linear interpolation at entropy 5.16 used in Section~\ref{sec:results} and Figure~\ref{fig:curves}: 42.1 / 52.4 / 58.8 for blocks 4 / 8 / 16. Checkpoints: \texttt{ELF-\{B,M,L\}-owt-torch}, \texttt{bd3lm-owt-block\_size\{4,8,16\}}, \texttt{mdlm-owt}, the \texttt{ar.ckpt} released with MDLM, and \texttt{gpt2}. ELF and MDLM run in bfloat16; BD3-LM uses its released FP32 backbone; GPT-2 small and our models use TF32. $^{a}$~whole canvas; $^{b}$~300 samples; $^{c}$~whole canvas under our judge: 38.2 / 5.28; $^{d}$~whole canvas under our judge: 12.8 / 5.28.}
\label{tab:baselines}
\label{tab:bd3sweep}
\begin{center}
\footnotesize\setlength{\tabcolsep}{4pt}
\begin{tabular}{llrrrr}
\toprule
& & \multicolumn{2}{c}{Published} & \multicolumn{2}{c}{Re-evaluated} \\
Model & Sampling & \genppl{} & Entropy & \genppl{} & Entropy \\
\midrule
ELF-B & SDE, 32 steps, guidance 3, $\gamma{=}1.5$ & 24.1 & 5.15 & 23.97 & 5.158 \\
ELF-B & SDE, 64 steps, guidance 3, $\gamma{=}1.0$ & --- & --- & 19.42 & 5.096 \\
ELF-M & SDE, 64 steps, guidance 3, $\gamma{=}1.0$ & 21.7 & 5.18 & 21.65 & 5.184 \\
ELF-L & SDE, 64 steps, guidance 3, $\gamma{=}1.0$ & 23.3 & 5.28 & 23.65 & 5.281 \\
Cosmos, $L{=}512$ & Euler SDE, 200 steps & --- & --- & 90.65 & 4.760 \\
BD3-LM, block 4 & first-hitting, $T{=}5000$, nucleus 0.9 & 25.7$^{a}$ & --- & 23.75 & 5.04 \\
BD3-LM, block 4 & same, nucleus 0.95 & --- & --- & 37.62 & 5.138 \\
BD3-LM, block 4 & same, nucleus 1.0 & --- & --- & 82.09 & 5.291 \\
BD3-LM, block 8 & same & 30.4$^{a}$ & --- & 28.00 & 5.02 \\
BD3-LM, block 8 & same, nucleus 0.95 & --- & --- & 45.09 & 5.137 \\
BD3-LM, block 8 & same, nucleus 1.0 & --- & --- & 98.58 & 5.256 \\
BD3-LM, block 16 & same & 33.4$^{a}$ & --- & 31.34 & 5.05 \\
BD3-LM, block 16 & same, nucleus 0.95 & --- & --- & 50.68 & 5.144 \\
BD3-LM, block 16 & same, nucleus 1.0 & --- & --- & 107.8 & 5.225 \\
MDLM & $T{=}1000$, nucleus 0.9 & 46.8$^{a,b}$ & --- & 35.4$^{c}$ & 4.90 \\
AR (MDLM release) & nucleus 0.9 & 14.1$^{a}$ & --- & 12.6$^{d}$ & 4.71 \\
GPT-2 small & $t{=}1.0$, 1023 new tokens & --- & --- & 12.1 & 5.12 \\
GPT-2 small & $t{=}1.1$, 1023 new tokens & --- & --- & 17.0 & 5.27 \\
GPT-2 small & $t{=}1.2$, 1023 new tokens & --- & --- & 23.8 & 5.40 \\
\bottomrule
\end{tabular}
\end{center}
\end{table}

BD3-LM's perplexity--entropy curve is steeper than ours (Appendix~\ref{app:metrics}), so its interpolated perplexity at entropy 5.16 lies well above its measured nucleus-0.9 points at entropy 5.02--5.05.

\paragraph{Cosmos.}
\label{app:cosmos}
Cosmos~\citep{meshchaninov2026cosmos} learns smooth BERT representations and reconstructs tokens in parallel. We evaluate its longest released OpenWebText checkpoint at its native 512 BERT tokens and 512 latents, without latent compression. Sampling uses 200 Euler SDE steps with the tan schedule ($d{=}5$), EMA weights, self-conditioning and a parallel greedy decoder: 279M generation parameters and 26.7 TFLOPs per native-length document. Denoising is FP32, with bfloat16 decoder autocast.

We sample 1000 documents (500 each at seeds 42 and 43, batch 32), truncate at the first SEP token, and apply the common GPT-2 Large judge, obtaining \genppl{} 90.65 and entropy 4.760. MAUVE uses the first 256 judge tokens of both generated texts and the same 1000 held-out references: MAUVE@256 is $0.677\pm0.034$ over five quantization seeds. Generation remains at the native length. Eager sampling takes 669.7\,ms per document, averaged over 28 full batches from two independent single-H100 runs, excluding each warmup and final partial batch.

We report this checkpoint only in the appendix because its tokenizer, training preprocessing and native length differ from the 1024-token setup. Its compute is for 512 BERT tokens, and its MAUVE@256 is not directly comparable to the main table's MAUVE@512.

\subsection{ELF-B step count and sampler settings}
\label{app:elfsteps}

ELF's released sampler has no temperature knob; step count, SDE noise re-injection scale $\gamma$ and self-conditioning guidance determine its operating point. Table~\ref{tab:elfsteps} evaluates 32, 40, 48 and 64 SDE steps under our judge. The 32-step released setting anchors our fixed-compute comparison. At that step count, tested noise scales ($\gamma\in\{0.5,1,2\}$) and guidance values (2 or 4 instead of 3) give at most about a 2\% improvement (23.46 at entropy 5.152 with guidance 4), comparable to sampling variation.

Increasing steps changes perplexity, entropy and repetition together: loop rate rises from 4\% at 32 steps to 13--15\% at 48--64. Table~\ref{tab:main} therefore also reports the 64-step released default, $\gamma{=}1.0$ (19.42 at entropy 5.096, loop rate 14.6\%). At 64 steps, $\gamma{=}1.5$ gives 17.57 at 5.067 (13.1\%), while $\gamma{=}2$ gives 15.5 at entropy 5.02 with 19\% looped documents.

At matched entropy, plain \ours{} is 0.2--3.5\% behind the 48--64-step points with $\gamma\le1.5$ (entropy 5.07--5.10; neighbours: $K{=}8$, $t{=}0.68/0.70/0.72$ in Figure~\ref{fig:sampling}(a)). The states-only rows and \ours{}-L stay ahead of those points; we claim no advantage over the 64-step $\gamma{=}2$ point.

\begin{table}[!htbp]
\caption{ELF-B (105M) with 32--64 SDE steps, 1000 documents, our judge; guidance 3 unless given in parentheses.}
\label{tab:elfsteps}
\begin{center}
\small
\begin{tabular}{rlrrr}
\toprule
Steps & $\gamma$ & \genppl{} & Entropy & Loop \\
\midrule
32 & 1.5 & 23.97 & 5.158 & 4.0\% \\
40 & 1.5 & 21.80 & 5.138 & 6.6\% \\
48 & 1.5 & 19.22 & 5.097 & 14.6\% \\
64 & 1.5 & 17.57 & 5.067 & 13.1\% \\
64 & 1.0 & 19.42 & 5.096 & 14.6\% \\
\midrule
32 & 0.5 & 39.40 & 5.296 & 1.6\% \\
32 & 1.0 & 28.71 & 5.203 & 2.7\% \\
32 & 2.0 & 26.90 & 5.148 & 3.6\% \\
32 & 1.5 (guidance 2) & 25.29 & 5.173 & 4.3\% \\
32 & 1.5 (guidance 4) & 23.46 & 5.152 & 5.0\% \\
64 & 2.0 & 15.52 & 5.024 & 19.0\% \\
\bottomrule
\end{tabular}
\end{center}
\end{table}

\subsection{A same-data autoregressive baseline}
\label{app:ar}

The same-data AR provides an autoregressive reference with our text preprocessing, generation-parameter count and an approximately matched training-token budget. It uses the block decoder's pre-tokenized OpenWebText (T5 tokenizer, 1024-token canvas, padding after end-of-text), trained for 103k steps at global batch 512: 50B non-padding tokens, within 7\% of \ours{}'s 54B.

The model has $d=1024$, 13 layers, 16 heads, a GELU MLP of width $4d$, rotary positions, tied input/output embeddings and no biases. Its 197M parameters match \ours{}'s decoder plus prior. It uses the optimizer and schedule of Table~\ref{tab:optim}; training took 10.8 hours on eight H100s (0.37 s per step, or 10.6 hours of steps, plus start-up, validation and checkpointing). Held-out negative log-likelihood is 2.635 nats per T5 token, versus 2.97 for \ours{}'s $K{=}8$ decoder teacher-forced on the true plan (Table~\ref{tab:blocklen}). Sampling uses a key--value cache and stops each document at its first end-of-text token.

\begin{table}[!htbp]
\caption{The same-data AR baseline under the paper's protocol (1000 documents, GPT-2 Large judge, per-document entropy, loop rate). Nucleus 0.86 and 0.88 were sampled in a later run than 0.85 and 0.90, hence the non-monotone entropies.}
\label{tab:ar}
\begin{center}
\small
\begin{tabular}{lrrr}
\toprule
Setting & \genppl{} & Entropy & Loop \\
\midrule
$t{=}0.80$ & 9.80 & 4.855 & 10.2\% \\
$t{=}0.85$ & 14.32 & 5.051 & 5.9\% \\
$t{=}0.88$ & 17.72 & 5.137 & 5.4\% \\
$t{=}0.90$ & 21.81 & 5.233 & 4.7\% \\
$t{=}1$, nucleus 0.85 & 12.90 & 5.082 & 4.1\% \\
$t{=}1$, nucleus 0.86 & 13.26 & 5.057 & 5.7\% \\
$t{=}1$, nucleus 0.88 & 15.37 & 5.151 & 3.6\% \\
$t{=}1$, nucleus 0.90 & 17.76 & 5.183 & 3.6\% \\
$t{=}1$, nucleus 0.95 & 27.74 & 5.337 & 3.5\% \\
$t{=}1$ & 67.73 & 5.569 & 1.7\% \\
\midrule
temperature sweep interpolated at entropy 5.16 & 18.6 & 5.16 & \\
nucleus sweep interpolated at entropy 5.16 & 16.0 & 5.16 & \\
\bottomrule
\end{tabular}
\end{center}
\end{table}

At entropy 5.16, the AR reaches 18.6 on the temperature sweep and 16.0 on the nucleus sweep (Table~\ref{tab:ar}; nucleus 0.88/0.90 brackets the target). Against 16.0, \ours{} at $K{=}8$ is 43\% behind with the plain sampler and 21\% behind with states-only nucleus sampling (19.4). \ours{}-L is 15\% behind with the plain sampler (18.4) and within 3\% with states-only sampling and the 4-gram ban (16.4), using 178 rather than 1024 dependent steps.

GPT-2 small remains stronger (13.2). At temperature 1, the same-data AR has entropy 5.57 versus GPT-2's 5.12 under common judge tokenization. The comparison constrains data and token-budget differences between our model and the same-data AR, but does not isolate which training or data differences explain GPT-2's advantage. FLOPs, serial depth and timing are in Table~\ref{tab:compute}; GPT-2 small starts its canvas with BOS and thus uses 1023 new-token evaluations.

\subsection{Sampling settings and the sweeps behind every interpolated number}
\label{app:sweeps}

\begin{figure}[!htbp]
\centering
\includegraphics[width=0.65\linewidth]{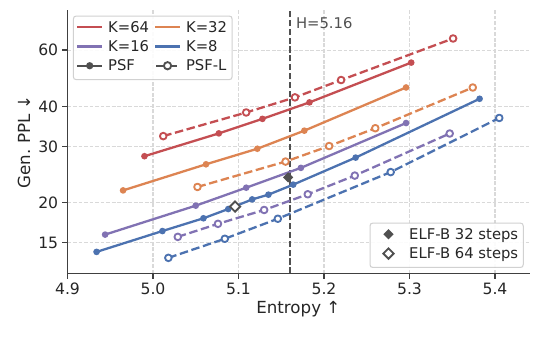}
\caption{\textbf{Perplexity--entropy sweeps by block count.} Plain-sampler curves for \ours{} and \ours{}-L at each $K$: color denotes block count, solid lines with filled circles denote \ours{}, and dashed lines with hollow circles denote \ours{}-L. Gray diamonds mark ELF-B at 32 and 64 steps. The vertical reference is entropy 5.16.}
\label{fig:sweeps}
\end{figure}

\emph{Selecting measured points.} Table~\ref{tab:main} reports actual sampled settings; Figure~\ref{fig:curves} uses interpolated comparisons. All our rows use guidance 1. Plain rows use the code-augmented prior and nucleus 0.95, with the temperature nearest entropy 5.16: 0.70 for \ours{} at $K{=}64,32,16$ and \ours{}-L at $K{=}8$; 0.75 for \ours{} at $K{=}8$; and 0.65 for \ours{}-L at $K{=}16$. The table retains, per size, these plain points and the lowest-perplexity states-only point near 5.16.

States-only rows (Appendix~\ref{app:retrain}; zero code targets and decoder inputs) use temperature 1: nucleus 0.55 with the 4-gram ban for \ours{}, and nucleus 0.60 without it for \ours{}-L. Other nearby candidates are \ours{}'s states-only $p{=}0.62$ (19.99 at 5.168, MAUVE 0.89) and \ours{}-L's $p{=}0.55$ with the ban (16.86 at 5.175, MAUVE 0.89). The latter's unbanned point is lower in perplexity (15.98 at 5.144) and is used in the main table. Baseline settings are in Appendix~\ref{app:baselines}; GPT-2 small uses temperature 1 and the same-data AR uses nucleus 0.88 (Appendix~\ref{app:ar}).

\begin{figure}[!htbp]
\centering
\includegraphics[width=\linewidth]{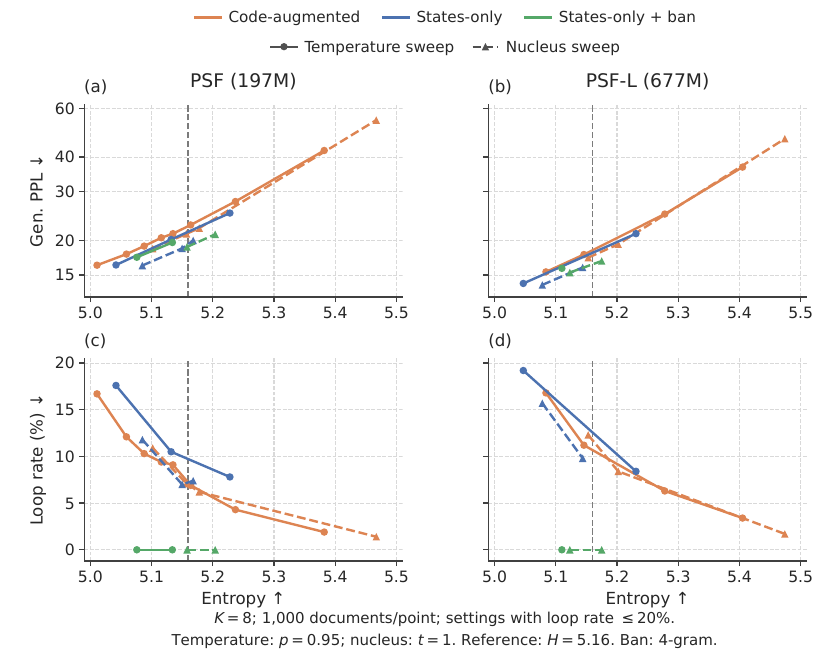}
\caption{\textbf{Sampling choices and repetition at $K{=}8$.} Left: \ours{}; right: \ours{}-L. Top: \genppl{}; bottom: loop rate. Temperature sweeps fix nucleus 0.95; nucleus sweeps fix temperature 1. Both rows show the same 36 settings with loop rate $\le20\%$ (1000 documents each); native $t{=}p{=}1$ settings are omitted. The vertical line marks entropy 5.16.}
\label{fig:sampling}
\end{figure}

\emph{Interpolation brackets.} The full sweep data support the comparisons at entropy 5.16. Figure~\ref{fig:sweeps} shows the plain multi-$K$ curves; Figure~\ref{fig:sampling} compares $K{=}8$ sampling choices and repetition for settings with loop rate at most 20\%. The accompanying \texttt{figure\_data/sampling/measurements.csv} retains all 77 settings, including native and higher-loop settings. All use guidance 1 except the $K{=}16$ guidance-4 setting (Appendix~\ref{app:seeds}); Appendix~\ref{app:norepeat} defines the 4-gram ban and Appendix~\ref{app:metrics} the loop rate.

Plain temperature brackets are 0.65/0.70 for \ours{} at $K=64,32,16$ and 0.70/0.80 at $K=8$; for \ours{}-L, they are 0.60/0.65 at $K=16$ and 0.70/0.80 at $K=8$. They give 39.2, 32.2, 24.9, 22.8, 20.3 and 18.4. States-only nucleus brackets are 0.60/0.62 for \ours{} (19.4), 0.55/0.60 with the ban (19.0), and 0.50/0.55 for \ours{}-L with the ban (16.4). Brackets span less than 0.08 nats except the $K=8$ plain brackets (0.15 for \ours{}, 0.13 for \ours{}-L). \ours{}-L's unbanned states-only sweep stops at 5.144: it does not bracket 5.16 and is not extrapolated.

\emph{Why figures and tables differ.} Figure~\ref{fig:curves} selects the lowest-perplexity sampler with a supported interpolation for each model size and block count: plain \ours{} at $K{=}16$ (24.9), and states-only sampling with the ban at $K{=}8$ (19.0 and 16.4). The banned setting is also \ours{}'s main-table and timed sampler; for \ours{}-L it supplies the bracket absent from the unbanned sweep. Panel (b) times \ours{}-L at nucleus 0.5 with the ban (45.38 / 32.15\,ms per document at batch 32 / 256).
For plain \ours{} at $K{=}8$, the tighter temperature pair 0.73/0.75 gives 22.5, while tables retain the higher 22.8 from 0.70/0.80. Figures~\ref{fig:sweeps} and \ref{fig:landing} join measured points and therefore cross 5.16 at 22.5. Table differences use 22.8; using 22.5 would make each ratio 1.3\% higher (e.g., $+4.8\%$ becomes $+6.2\%$).

\emph{Native distributions.} Without temperature sharpening or a nucleus cut ($t{=}1$, $p{=}1$), the native settings in the full sweep data have entropy 5.6--5.7 and perplexity 82--122; these settings are omitted from Figure~\ref{fig:sampling}. At held-out entropy 5.41 (between the $t{=}0.9$ and nucleus-0.9 points), \ours{} is about 45 and \ours{}-L about 37, versus 16.9 for held-out text. The headline comparisons thus concern sharpened samplers in the 5.0--5.2 entropy band.

\emph{Extended landing-target and baseline curves.}
The additional points in Figure~\ref{fig:landing}(b) extend the temperature sweeps to $t{=}0.925$--$0.95$ at nucleus 0.95. Every marker is measured on 1000 documents, except the 500-document switched-off $K{=}64$ probe. Figure~\ref{fig:landing}(a) uses ELF-B at 32 SDE steps with $\gamma\in\{0,0.25,0.5,1,1.5,2\}$ and at 64 steps with $\gamma\in\{0.5,1,1.5,2\}$, both at guidance 3; BD3-LM uses block size 4 and nucleus $p\in\{0.8,0.85,0.9,0.95,1\}$. The hollow dashed extensions fix 32-step ELF-B's $\gamma{=}0$ and reduce self-conditioning guidance from 3 through 2, 1.5 and 1.25 to 1; for BD3-LM they fix $p{=}1$ and increase temperature from 1 to 1.025, 1.05 and 1.075. They reach entropy 5.516 and 5.525, respectively. Their newly sampled anchors use different batch and random-number layouts from the original sweeps, so the anchors need not coincide. BD3-LM retains its official low-entropy rejection rule. All points use the same judge and first-document protocol, with no whole-canvas scores or extrapolations.

The GPT-2 small reference curve in Figure~\ref{fig:landing}(a) uses top-$k{=}50$, nucleus $p{=}1$, and temperatures $t\in\{1,1.1,1.2\}$, with no 4-gram ban. Each setting generates 1000 documents of 1023 new tokens after BOS using FP32 model weights.

\subsection{Seeds and variance}
\label{app:seeds}

Table~\ref{tab:seeds} measures variation across fresh sampling seeds at fixed trained checkpoints, not uncertainty across training runs. Each seed uses 1000 documents and, where indicated, its own bracketing points for interpolation at entropy 5.16. For our settings, standard deviations are 0.04--0.4 perplexity points and about 0.005 nats of entropy; for the baselines' headline points they are 0.2--0.8 points and up to 0.017 nats. Differences of about 1--2\% are therefore comparable to sampling variation in these checks. These are not error estimates for every cell: the table lists all settings sampled with multiple seeds, and other cells are single sampling runs.

\begin{table}[!htbp]
\caption{Seed-to-seed variance of the interpolated or measured generative perplexity (1000 documents per seed, GPT-2 Large judge). ``Reference'' uses the original evaluation sweep at the entropy specified in each row.}
\label{tab:seeds}
\begin{center}
\footnotesize\setlength{\tabcolsep}{3pt}
\begin{tabular}{l>{\raggedright\arraybackslash}p{2.9cm}l>{\raggedright\arraybackslash}p{3.0cm}l}
\toprule
Model & Setting & Seeds & Mean $\pm$ sd & Reference \\
\midrule
\ours{} $K{=}8$ & plain, at entropy 5.16 & 22.5 / 22.5 / 23.2 & $22.7 \pm 0.4$ & 22.8 \\
\ours{}-L $K{=}8$ & plain, at entropy 5.16 & 18.4 / 18.6 / 18.6 & $18.5 \pm 0.1$ & 18.4 \\
\ours{}-L $K{=}16$ & plain, at entropy 5.16 & three seeds & $20.5 \pm 0.2$ & 20.3 \\
\ours{} $K{=}16$ & $t{=}0.70$, guidance 4 (measured) & four seeds & $24.81 \pm 0.07$ (entropy $5.166 \pm 0.003$) & 24.80 \\
\ours{} $K{=}8$ & nucleus, zero codes at inference, at 5.16 & 19.1 / 19.4 / 19.5 & $19.3 \pm 0.2$ & 19.3 (Figure~\ref{fig:codes}) \\
\ours{} $K{=}8$ & states-only, nucleus 0.6, at 5.16 & 19.4 / 19.4 / 19.3 & $19.4 \pm 0.1$ & 19.4 \\
\ours{} $K{=}8$ & states-only, nucleus 0.55, ban, at 5.16 & 19.0 / 19.1 / 19.3 & $19.2 \pm 0.1$ & 19.0 \\
\ours{}-L $K{=}8$ & states-only, nucleus 0.6 (measured) & 16.0 / 15.9 / 15.8 & $15.9 \pm 0.1$ (entropy $5.142 \pm 0.004$) & 16.0 \\
\ours{}-L $K{=}8$ & states-only, nucleus 0.5/0.55, ban, at 5.16 & 16.4 / 16.3 / 16.4 & $16.4 \pm 0.04$ & 16.4 \\
\midrule
ELF-B, 32 steps & $\gamma{=}1.5$, at entropy 5.16 & 24.0 / 24.2 / 23.5 & $23.9 \pm 0.4$ & 24.0 \\
ELF-B, 64 steps & $\gamma{=}1.0$ (measured) & 19.42 / 18.66 / 18.40 & $18.8 \pm 0.5$ (entropy $5.078 \pm 0.017$) & 19.42 \\
BD3-LM, block 4 & nucleus 0.9 (measured) & 23.75 / 23.35 / 23.78 & $23.6 \pm 0.2$ (entropy $5.025 \pm 0.013$) & 23.75 \\
BD3-LM, block 4 & nucleus 0.95 (measured) & 37.62 / 37.64 / 36.26 & $37.2 \pm 0.8$ (entropy $5.135 \pm 0.006$) & 37.62 \\
\bottomrule
\end{tabular}
\end{center}
\end{table}

\FloatBarrier
\section{Additional ablations}
\label{app:addabl}

These ablations separate the effects of training, inference-time interventions and sampling choices. Unless otherwise specified, we use \ours{} with $K{=}8$, the plain sampler and the prior's codes, and read generative perplexity at entropy 5.16 (1000 documents, GPT-2 Large judge).

\subsection{The landing target and the training recipe}
\label{app:landing}
\label{app:seam}

\begin{table}[!htbp]
\caption{Recipe ablations at $K{=}64$, compared at matched entropy. Added components are measured against the model without them when introduced; removal from the final model is marked separately (seed sd 0.04--0.4, Appendix~\ref{app:seeds}).}
\label{tab:ablation}
\begin{center}
\footnotesize\setlength{\tabcolsep}{3pt}
\begin{tabular}{>{\raggedright\arraybackslash}p{4.2cm}>{\raggedright\arraybackslash}p{4.9cm}>{\raggedright\arraybackslash}p{4.2cm}}
\toprule
Comparison & $\Delta$\genppl{} at matched entropy & Diagnostic \\
\midrule
Landing target added & $-25\%$ & seam perplexity 9241 $\to$ 958 \\
No landing target (final model) & $+13$ to $+15\%$ & codes kept \\
Realization code added & $-7$ to $-11\%$ (45.4 $\to$ 40.6 at 5.16) & greedy repeats 55\% $\to$ 14\% \\
Code lookahead $\vr_{k+1}$ added & $-2$ to $-3\%$ & block-final tokens $-6$ to $-7\%$ \\
Self-realized prefixes added & $-6.4\%$ (69.8 $\to$ 65.3 at 5.34) & true plan $+30\%$ without label \\
Source label $\lambda_k$ added & $-2.1\%$ (65.3 $\to$ 63.9 at 5.34) & restores true-plan decode \\
Prior: mean codes vs.\ sampled & $-4.5\%$ & --- \\
Prior: logit-normal $t$ vs.\ uniform & $-5$ to $-6\%$ & --- \\
\bottomrule
\end{tabular}
\end{center}
\end{table}

Table~\ref{tab:ablation} records the development of the recipe at $K{=}64$, before it was carried over to other block lengths. The additions were tested at different training stages, so their gains are not additive contributions or a ranking at a common checkpoint. The landing target gives a large reduction when introduced, and removing it from the final model still costs 13--15\%. The training interventions also interact: self-realized prefixes alone worsen true-plan decode, while the source label restores it and brings sampled-plan perplexity close to true-plan decoding. The realization code reduces greedy repetition as well as perplexity; the two prior-side choices each give about 5\%. These results support the full training recipe, whose inference-time components are tested separately below.

\paragraph{Landing-target comparisons.}
Figure~\ref{fig:landing}(b) separates retraining without the target from switching it off at inference, which replaces it by the trained ``no target'' vector at every position, as for the last block. Its $K{=}64$ checkpoint pair is the final-model removal in Table~\ref{tab:ablation}. It uses a uniform-time prior; the logit-normal prior in Table~\ref{tab:blocklen} lowers perplexity by 5--6\%, explaining why the figure's curves lie above that table's $K{=}64$ value. Without the target, $K{=}64$ documents are also shorter. Their lower per-document entropy under the 1024-token protocol can contribute to the matched-entropy gap (Appendix~\ref{app:length}). Table~\ref{tab:seam} instead compares positions at a fixed setting, averaging blocks 2--64 of 500 documents at $t{=}0.65$ and nucleus 0.95, without this entropy-matching effect. At $K{=}8$ the lengths are nearly equal. Appendix~\ref{app:sweeps} records the extended curve settings.

At $K{=}8$, switching off the target gives 23.3 against 22.5 with the nearest brackets as plotted ($+3.3\%$; $+3.2\%$ with the 0.70/0.80 pairs on both sides). Retraining gives 23.9 / 22.2 with the plain / nucleus sampler: the figure's like-for-like gaps are $+6.2\%$ / $+4.7\%$, or $+5.2\%$ for plain with the 0.70/0.80 pairs. Table~\ref{tab:retrain} reports $+4.8\%$ for plain against its standard 22.8 baseline, rather than the figure's 22.5 (Appendix~\ref{app:sweeps}). The unplotted inference intervention on \ours{}-L gives 19.2 against 18.4 ($+4.5\%$).

\paragraph{The seam with the target.}
Even with the landing target, the seams remain the most expensive positions. At $K{=}8$ (entropy 5.16; blocks 2 to $K$ of 500 documents) the judge's perplexity at the first token of a block is 332 against 18 in mid-block, and rises again to 200 at the last position, where the block lands on the next block's first token; at $K{=}16$ (entropy 5.11) the first token costs 145 against 17. Halving $K$ makes boundaries less frequent, consistent with fewer, longer blocks helping generation (Section~\ref{sec:blocklen}).

\paragraph{Lookahead and source labels at inference.}
On the final $K{=}8$ models, zeroing the lookahead code $\vr_{k+1}$ gives 22.4 against 22.8, and setting the source label to ``original state'' ($\lambda_k{=}1$) gives 24.3 against 24.7 at entropy 5.191, but the loop rate rises to 9.3\% and 9.9\% (against 6.9\% and 9\%); for \ours{}-L the lookahead gives 17.2 against 17.1 at entropy 5.130. Perplexity changes little in these inference checks, even where looping increases. Together with the training ablations, they support retaining the landing target, lookahead code and source label at their trained settings by default.

\subsection{Realization codes, the states-only prior and content states}
\label{app:retrain}
\label{app:infabl}

\begin{figure}[!htbp]
\centering
\includegraphics[width=\linewidth]{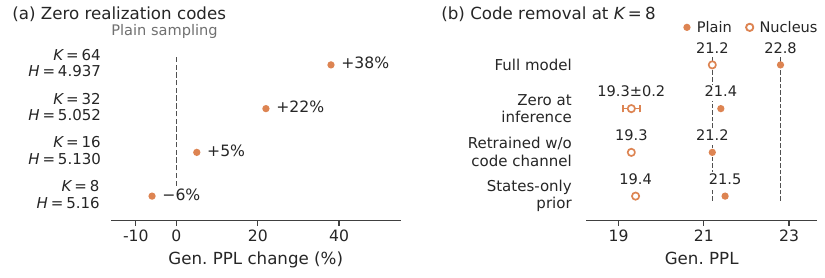}
\caption{\textbf{Realization-code ablations.} (a) Change in generative perplexity after replacing the prior's codes with zeros, with entropy matched within each labeled $K$ row. (b) $K{=}8$, entropy 5.16; dashed guides mark the full-model references. $\pm$: standard deviation over three seeds. Full/zero $K{=}8$ results also appear in Table~\ref{tab:content}.}
\label{fig:codes}
\end{figure}

\begin{table}[!htbp]
\caption{Additional realization-code and retraining controls ($K{=}8$, entropy 5.16 unless stated). ``vs.'' uses the full model at matched sampler/entropy, with its PPL/loop in parentheses; content states use the corresponding predictive-state control. nuc.\ = nucleus; ban = 4-gram ban; $^\star$: Table~\ref{tab:content}; $^\ddagger$: extrapolated by 0.015 nats (zero codes) / 0.025 nats (true plan), Appendix~\ref{app:content}.}
\label{tab:retrain}
\label{tab:infabl-codes}
\label{tab:infabl-k}
\centering
\footnotesize\setlength{\tabcolsep}{2pt}
\begin{tabular}{l>{\raggedright\arraybackslash}p{2.2cm}r@{\hspace{6.5pt}}r@{\hspace{6.5pt}}r}
\toprule
Variant & Sampler & \genppl{} & vs.\ control & Loop \\
\midrule
\multicolumn{5}{l}{\emph{\ours{} $K{=}8$}} \\
shuffled codes & plain & 23.4 at 5.174 & $-1\%$ (23.7) & 9.1\% (7\%) \\
codes $\sim \mathcal{N}(0, I)$ & plain & 19.95 at 5.164 & $-13\%$ (23.0) & 19.3\% \\
true plan, true $\to$ zero codes & $t{=}0.65$ & 19.9 at 5.168 & $-19\%$ (24.5) & 18.1\% (14\%) \\
retrained w/o code channel & nuc., ban & 19.0 & $-10\%$ & 0\% \\
states-only prior & nuc. 0.55, ban & 19.0 & $-10\%$ & 0\% \\
retrained w/o landing target & plain / nuc. & 23.9 / 22.2 & $+4.8\%$ / $+4.7\%$ & 7.6\% / 8.2\% \\
content states $\vc_k$$^\star$ & plain / nuc. 0.55--0.60 & 24.7 / 23.2 & $+8\%$ / $+9\%$ & 7.3\% / 8.5\% \\
content states $\vc_k$$^\star$ & zero codes, plain / nuc. & 24.0$^\ddagger$ / 22.4 & $+12\%$ / $+16\%$ & 7.5\% / 5.0\% \\
content states $\vc_k$$^\star$ & true-plan decode & 26.9$^\ddagger$ & $+12\%$ & \\
\midrule
\multicolumn{5}{l}{\emph{\ours{}-L $K{=}8$}} \\
full model (prior's codes) & plain / nuc. 0.55 & 18.4 / 17.6 & & 11.2\% / 12.3\% \\
zero codes at inference & plain / nuc. & 18.0 / 15.8 at 5.140 & $-2\%$ / $-6\%$ (16.8) & 11\% / 9.3\% (13\%) \\
states-only prior & nuc. 0.6 & 16.0 at 5.144 & $-6\%$ (17.0) & 9.8\% \\
states-only prior & nuc. 0.5/0.55, ban & 16.4 & $-7\%$ & 0\% \\
\bottomrule
\end{tabular}
\end{table}

Figure~\ref{fig:codes} and Table~\ref{tab:retrain} separate inference-time dependence on codes from their role during training. At inference we replace the prior's codes by zeros, fresh $\mathcal{N}(0, I)$ draws, another document's codes from the batch (``shuffled'') or scaled copies. For retraining we repeat the last chain stage with the change: 6k self-realized-prefix steps from the $K{=}16$ decoder, followed by 30k prior steps (Appendix~\ref{app:hparams}). Both sizes retain the encoder trunk frozen after their base stage (Table~\ref{tab:stages}).

\paragraph{Codes at inference.}
At $K{=}8$, the prior's codes do not improve generative perplexity at matched entropy: shuffling has little effect, and zero codes are 6\% (plain) to 9\% (nucleus) better, with higher loop rates. Fresh $\mathcal{N}(0, I)$ codes are 13\% better but raise looping to 19.3\%, while scaling codes by 1.5 is 10\% worse. The sign reverses with shorter blocks: zeroing codes costs 38\%, 22\% and 5\% at $K{=}64$, 32 and 16.

Under teacher forcing, posterior-mean codes lower decoder NLL at every block length, by 0.94, 0.91, 0.77 and 0.53 nats per token at $K{=}64$, 32, 16 and 8 (\ours{}-L: 1.14, 1.08, 0.91 and 0.57). Codes inferred by the prior for the true states recover 80--90\% of this gain. The code is a linear read-out of the next predictive state with a budget of about one nat per token; its decreasing NLL benefit is consistent with longer blocks providing more local history. In generation, codes reduce entropy along with perplexity at a fixed temperature, so that benefit need not persist at matched entropy. The $K{=}8$ reversal is also present with true plans: zero instead of true codes gives 19.9 against 24.5 at entropy 5.168 ($-19\%$), so it cannot be explained solely by prior prediction errors. The prior's codes have similar marginal scales to the encoder's posterior means (standard deviation 1.44 against 1.59 per dimension, norm 11.7 against 13.1).

The perplexity gain also has a cost in the paired generation probes. On the same sampled plans at $K{=}8$, zero codes lower perplexity by 6.3\% (plain) and 9.7\% (nucleus) at matched entropy, but MAUVE@512 falls by 0.052 / 0.051 and the lexical topic-switch rate rises from 6.9\% to 11.8\% and from 6.8\% to 13.1\%. Thus the lowest-perplexity choice need not best preserve the sample distribution or topic persistence.

\paragraph{Retraining and the states-only prior.}
A decoder retrained without the code channel reaches the zero-code generation numbers in Figure~\ref{fig:codes}, but its true-plan decode is 18\% worse than the code-trained decoder's with zero codes (21.4 against 18.1 at entropy 5.13). This supports a training benefit under the shared adaptation protocol, despite the inference-time perplexity result. We retain that code-trained decoder and train a \emph{states-only prior} with zero code targets, continuing the $K{=}8$ prior for 30k steps. Its generated codes come out at zero and it reproduces the zero-code numbers for both sizes without changing the architecture. It supplies the $K{=}8$ states-only rows of Table~\ref{tab:main}; at $K\ge16$ we retain generated codes.

\paragraph{Predictive against content states.}
\label{app:content}
The same adaptation recipe tests whether a block should read a state of its prefix or of the block itself. The control replaces the predictive state $\vs_k = \vc_{k-1}$ by the inclusive state $\vc_k = f_\psi(x^{(\le k)})$, a content latent as in segment-level latent diffusion. It retains the frozen encoder trunk, code channel, 6k self-realized-prefix steps from the $K{=}16$ checkpoint and 30k prior steps. Both variants inherit a decoder trained on predictive states, so this tests the indexing choice under shared adaptation, with an initialization that may favor predictive states.

The content control has 8--16\% higher perplexity across sampling settings and 12\% higher true-plan perplexity (Tables~\ref{tab:content} and \ref{tab:retrain}). The latter gap persists when prior sampling is bypassed, suggesting that it depends partly on how the decoder uses the representation. Loop rates do not follow the same ordering: they are similar with the plain sampler (7.3\% against 6.9\% at $t{=}0.75$), and lower for content with zero codes (7.5\% / 5.0\% against 9.7\% / 7.7\%). The 4-gram ban changes matched-entropy perplexity by only $-2\%$ to $+1\%$ (Appendix~\ref{app:norepeat}). Two content readings require small extrapolations: 0.015 nats for zero codes with plain sampling and 0.025 nats for true-plan decode.

\paragraph{What the sampled plan controls.}
\label{app:planprobe}
Four probes on \ours{} at $K{=}8$ (1000 documents each; the second also on \ours{}-L) examine plan use in prediction, generation and representation space. First, under teacher forcing, replacing a document's predictive states by another document's, by the dataset-mean state or by its own states with blocks 2--8 permuted costs $+8.0\%$, $+7.1\%$ and $+2.6\%$ in NLL ($+9.5\%$, $+5.4\%$ and $+1.8\%$ with zero codes).

The generation probe replaces sampled plans by their position-wise mean, with zero codes and the same block-decoder noise. At fixed sampler settings, this raises perplexity by 7.0--8.7\% for \ours{} (plain, nucleus, states-only, states-only with the ban), and for \ours{}-L by 7.4--9.6\% with nucleus sampling and 0.7--1.5\% with the ban. We use fixed settings for this comparison because the intervention also changes length and entropy: \ours{}'s mean-plan documents are 11--13\% shorter and switch topics more often. Its matched-entropy perplexity differences range from $-16\%$ to $+22\%$ across models and scoring windows, making them difficult to interpret as a plan effect alone.

The mean plan also lowers MAUVE@512 from 0.86--0.88 to 0.71--0.75 and raises the lexical topic-switch rate from 11.8--13.7\% to 30--47\% (Appendix~\ref{app:topic}). These summaries, and those in the code probe above, are interpolated to entropy 5.16. For \ours{}-L, MAUVE falls from 0.86--0.91 to 0.61--0.71 (read at entropy 5.18--5.26 without interpolation), and switch rates rise from 10\% to 48--51\% (real text: 3.6\%). A fixed-setting check gives paired MAUVE scores of 0.866 versus 0.754 for sampled versus mean states (plain sampler, $t=0.75$, $p=0.95$, zero codes). MAUVE reflects a distributional change that includes sharing one plan across documents; lexical switches track within-document topic persistence.

Finally, re-encoded outputs are closer to their own plan than another sample's in 98\% of documents: mean cosine is 0.40--0.41 versus 0.13--0.14 for blocks 2--8, except block 2 (0.11, as for a random plan). Sampled plans also cover encoded ones, with $k$-nearest-neighbour precision 0.78--0.79 and recall 0.87--0.88 ($k{=}5$). Together, the probes support the plan's role in organizing content.

\subsection{Block length, scale and training budget}
\label{app:blocklen}

\begin{table}[!htbp]
\caption{Reported numbers behind Figure~\ref{fig:ablations}: block length and scale with the recipe held fixed (plain sampler, the prior's codes at every $K$). Teacher-forced NLL = held-out cross-entropy of the block decoder in nats per token; perplexities at entropy 5.16; true-plan decode as defined in Appendix~\ref{app:metrics}; loop rate at the setting closest to 5.16.}
\label{tab:blocklen}
\label{tab:scaling}
\begin{center}
\small\setlength{\tabcolsep}{4pt}
\begin{tabular}{rr*{8}{>{\centering\arraybackslash}p{1.1cm}}}
\toprule
& & \multicolumn{2}{c}{\shortstack{Teacher-forced\\NLL}} & \multicolumn{2}{c}{\shortstack{\genppl{}\\\iso}} & \multicolumn{2}{c}{\shortstack{True-plan\\decode \iso}} & \multicolumn{2}{c}{Loop} \\
$K$ & $\ell$ & \ours{} & \ours{}-L & \ours{} & \ours{}-L & \ours{} & \ours{}-L & \ours{} & \ours{}-L \\
\midrule
64 & 16 & 2.63 & 2.45 & 39.2 & 42.2 & 40.3 & 35.1 & 1.2\% & 7\% \\
32 & 32 & 2.78 & 2.57 & 32.2 & 27.1 & 31.0 & 25.6 & 2.8\% & 9\% \\
16 & 64 & 2.89 & 2.69 & 24.9 & 20.3 & 26.5 & 21.2 & 7.6\% & 10\% \\
8 & 128 & 2.97 & 2.77 & 22.8 & 18.4 & 24.0 & 19.1 & 6.9\% & 11.2\% \\
\bottomrule
\end{tabular}
\end{center}
\end{table}

Table~\ref{tab:blocklen} follows the training chain of Section~\ref{sec:training}; it does not isolate block length at equal training budget. Both sizes retain their encoder trunk frozen after the base stage. Along this chain, each halving of $K$ lowers perplexity, the prior's cost falls from 631 to 138 GFLOPs, and the block decoder's stays near 172. For \ours{}, sampled-plan generation is at most about 4\% worse than true-plan decode and 4--6\% better at $K \le 16$; the plan probes above test content organization separately. \ours{}-L, with $d = 1024$ and a 308M prior, has 7--8\% lower teacher-forced cross-entropy, better true-plan decode at every $K$, and 16\%, 18\% and 19\% lower generative perplexity at $K = 32$, 16 and 8. We use $K{=}8$ for both sizes and examine the role of training below.

\paragraph{Direct fine-tuning against the chain.}
\label{app:direct}
Fine-tuning the pre-chain $K{=}64$ block decoder directly at $K{=}8$ with the last stage's 6k steps reaches 24.5 against the chain's $22.7 \pm 0.4$ (three seeds). Continued to the chain's total of 20k block-decoder steps, with its prior resumed for 30k more, it reaches 22.9 (chain 22.8), 21.6 at entropy 5.160 with nucleus 0.62 (chain 21.2) and a true-plan decode of 23.5 at 5.153 (chain 23.7 at 5.155). It loops slightly more (8.9\% against 6.9\% at $t{=}0.75$). The initial gap therefore closes with additional decoder and prior training. The direct route's 20k steps took 37 hours on eight H100s; we retain the chain, which spends fewer steps on expensive $K{=}8$ rollouts and yields the $K{=}64$, 32 and 16 models as by-products.

\paragraph{Continued training of \ours{}-L.}
\label{app:bigcont}
At $K{=}8$, continuing \ours{}-L's block decoder for 6k and 12k more steps (+6\% and +13\% tokens) improves perplexity from 18.4 to 18.2 and 18.0 (seed sd 0.1), a modest gain over the tested extension. The $K{=}64$ exception in Table~\ref{tab:blocklen} is more sensitive to training: \ours{}-L initially generates worse than \ours{} (42.2 against 39.2) despite better true-plan decode (35.1 against 40.3). Another 30k prior steps leave it at 42.6, while 10k more self-realized-prefix decoder steps and 30k prior steps bring it to 36.4 at entropy 5.16 (between 36.09 at 5.156 and 41.55 at 5.223), 7\% ahead of \ours{}, with true-plan decode about 33.5 and looping at 8.3\%. Its $K{=}64$ stage had seen 22k self-realized-prefix steps at batch 192 against \ours{}'s 20k at batch 512. The improvement beyond prior-only continuation points to decoder adaptation as a contributing factor at this block length. The continued model is the star in Figure~\ref{fig:ablations}; Table~\ref{tab:blocklen} retains the original model for comparability along the chain.

\subsection{Prior steps and refinement}
\label{app:sampling}
\label{app:nfe}

Changing the prior step count also shifts entropy at a fixed sampler setting, so we re-read the sweeps at entropy 5.16 with 32 and 64 instead of 50 steps. Every headline $K{=}8$ value whose sweep brackets 5.16 changes by about 1\%, with no consistent direction. For 50 $\to$ 32 / 64 steps, the readings are: plain 22.5 $\to$ 22.6 / 22.8 with the tightest pair (22.8 $\to$ 22.9 / 23.0 with the 0.70/0.80 pair of Appendix~\ref{app:sweeps}); states-only nucleus 19.4 $\to$ 19.2 / 19.2, or 19.0 $\to$ 19.1 / 19.0 with the ban; \ours{}-L plain 18.4 $\to$ 18.3 / 18.5, and states-only with the ban 16.4 $\to$ 16.2 / 16.4. The re-read $K{=}8$ ablations (Table~\ref{tab:content} and Figure~\ref{fig:landing}(b)) move by at most 1.8\%, except inference-time removal of the landing target at 64 steps ($+2.9\%$).

The shorter-block models are more sensitive. At $K{\ge}16$, 32 steps cost 1--3.4\% (\ours{}-L $K{=}32$: 6.5\%), while 64 steps gain at most 1\% (\ours{}-L $K{=}32$: 3.1\%). Earlier runs with 20 and 12 steps gave $+0.3\%$ and $-0.2\%$ at $K{=}8$ (nucleus 0.6), and $+2.4\%$ and $+5.2\%$ at $K{=}16$ (guidance 4).

A separate fixed-setting MAUVE@512 check gives 0.926 / 0.914 for the plain sampler and 0.847 / 0.834 for states-only with the ban, at 50 / 32 steps. It uses 1000 documents per setting and five quantization seeds on saved generations, rather than matched-entropy estimates. This check also finds small changes at $K{=}8$.

\paragraph{Runtime and comparison with ELF-B.}
At batch 32, the $K{=}8$ plain sampler takes 25.9 / 27.2 / 28.2\,ms per document at 32 / 50 / 64 steps in one session (Table~\ref{tab:main}: 27.3\,ms). Thus reducing prior steps gives a smaller reduction in end-to-end time. At 32 steps and entropy 5.16, \ours{} has lower perplexity than ELF-B (22.9 versus 24.0 with the 0.70/0.80 pair, 22.6 with the tightest; states-only 19.2 and 19.1), while ELF-B is faster at batch 32 (20.9 versus 25.9\,ms). At 64 steps, \ours{} is faster (28.2 versus 40.9\,ms); read at ELF-B's entropy of 5.096, its plain sampler is close (19.6 versus 19.42) and the states-only rows have about 15\% lower perplexity. We retain 50 steps as a common setting for our models in all tables.

\paragraph{Refinement.}
\label{app:refine}
The optional refinement round of Section~\ref{sec:generation} rewrites every block; the reference setting uses guidance 1, nucleus 0.95 and temperature 0.8 in both passes. Sweeping both temperatures to compare at matched entropy, one round is 4--10\% worse at every block length and size (\ours{} $K{=}8$, plain, 0.80 / 0.80: 22.8 at entropy 5.137 against 21.3 one-shot), and two rounds are 15\% worse. At a fixed temperature, refinement also raises looping (10\% to 15\% at $t{=}0.70$, 4.3\% to 6.5\% at $t{=}0.80$). Confidence-ordered partial commit, accepting half the blocks per round, brings perplexity within $\pm 2.5\%$ of one-shot. These tested refinement settings provide no consistent improvement, so we use one-shot generation.

\section{GPU time and throughput}
\label{app:batch}

\paragraph{Timing protocol.}
All runs use one idle H100 80GB after warm-up, excluding model loading, graph capture and text scoring. Times are per document (1024 tokens unless noted), with batched time divided by batch size. The current \ours{} and same-data AR measurements use synchronized wall-clock time, averaged over 5 runs, with the evaluated sampler. For \ours{}, CUDA graphs cover the 50-step prior and key--value-cached block decoder in TF32; temperature, nucleus selection and the 4-gram ban run between graph replays. The same-data AR uses a CUDA graph with a static key--value cache. Table~\ref{tab:main} reports batch-32 time per document.

GPT-2 small (CUDA graph, TF32), ELF-B/M/L (\texttt{torch.compile}, bfloat16), BD3-LM and MDLM (eager) come from earlier measurements on the same GPU model and harness family. Their batch-1 entries use the earlier harness's median wall-clock statistic with launch overhead removed, so their aggregation and overhead treatment differ from the current measurements. A batch-32 cross-check gives 26.6\,ms earlier versus 26.66\,ms now for greedy \ours{} at $K{=}8$. The table identifies eager implementations and the separate batch-scaling run.

\begin{table}[!htbp]
\caption{Cost of one 1024-token document on one H100. FLOPs from PyTorch's operator counter over the sampling code (batch 1; analytic for the same-data AR); train FLOPs $=6ND$ (ours cumulative along the chain, rollouts and prior included; ELF-B 46B tokens, assumed for ELF-M/L; BD3-LM/MDLM 524B). Timing protocol as in the paragraph above. Batch 256 for ELF-B and GPT-2 small comes from a separate batch-scaling run (ELF-B at batch 32 in that run: 46.7 documents per second).}
\label{tab:compute}
\begin{center}
\footnotesize\setlength{\tabcolsep}{2.5pt}
\begin{tabular}{@{}lrrrrrr@{}}
\toprule
Model (sampler) & Train FLOPs & FLOPs/doc & Serial & Latency & \multicolumn{2}{c}{Throughput (doc/s)} \\
 & ($\times 10^{19}$) & (TFLOPs) & depth & (batch 1, s) & batch 32 & batch 256 \\
\midrule
\ours{} $K{=}64$ ($t{=}0.7$) & 4.2 & 0.80 & 66 & 0.14 & 62.6 & 71.4 \\
\ours{} $K{=}32$ ($t{=}0.7$) & 4.6 & 0.52 & 82 & 0.18 & 70.2 & 92.3 \\
\ours{} $K{=}16$ ($t{=}0.7$) & 4.9 & 0.38 & 114 & 0.26 & 60.4 & 93.7 \\
\ours{} $K{=}8$, plain ($t{=}0.75$) & 5.2 & 0.31 & 178 & 0.42 & 36.7 & 62.9 \\
\ours{} $K{=}8$, states-only ($p{=}0.55$, ban) & 5.2 & 0.31 & 178 & 0.45 & 30.8 & 48.9 \\
\ours{}-L $K{=}16$ ($t{=}0.65$) & 5.2 & 1.35 & 114 & 0.38 & 29.0 & 38.1 \\
\ours{}-L $K{=}8$, plain ($t{=}0.7$) & 5.6 & 1.10 & 178 & 0.57 & 24.9 & 36.3 \\
\ours{}-L $K{=}8$, states-only ($p{=}0.60$) & 5.6 & 1.10 & 178 & 0.59 & 22.2 & 31.3 \\
\midrule
GPT-2 small & --- & 0.27 & 1023 & 2.41 & 7.8 & 10.9 \\
AR, same data (nucleus 0.88) & 6.4 & 0.43 & 1024 & 2.93 & 4.6 & 7.4 \\
\midrule
ELF-B, 32 / 64 steps & 3.2 & 7.2 / 14.2 & 33 / 65 & 0.13 / 0.26 & 47.8 / 24.5 & 49.8 / 25.2 \\
ELF-M, 32 / 64 steps & 9.6 & 26.0 / 51.2 & 33 / 65 & 0.26 / 0.50 & 12.3 / 6.2 & --- \\
ELF-L, 32 / 64 steps & 18 & 48.9 / 96.4 & 33 / 65 & 1.23 / 2.40 & 2.8 / 1.4 & --- \\
BD3-LM, block 4 & 53 & 17.9 & 1023 & 18.9$^\dagger$ & 0.71$^\dagger$ & --- \\
BD3-LM, block 8 & 53 & 17.3 & 1023 & 16.2$^\dagger$ & 0.73$^\dagger$ & --- \\
BD3-LM, block 16 & 53 & 18.7 & 1023 & 15.5$^\dagger$ & 0.68$^\dagger$ & --- \\
MDLM, $T{=}1000$ & 53 & $\approx$347 & 1000 & 34.4$^\dagger$ & 0.06$^\dagger$ & --- \\
\bottomrule
\multicolumn{7}{@{}p{0.97\textwidth}@{}}{\footnotesize $^\dagger$Eager sampler (no CUDA-graph step; BD3-LM/MDLM batch 32 via a row-wise re-implementation).} \\
\end{tabular}
\end{center}
\end{table}

\paragraph{Throughput and batch size.}
Lower FLOPs coexist with greater serial depth: at $K{=}8$, \ours{} uses 0.31 TFLOPs and 178 dependent network evaluations, versus 7.2 TFLOPs and 33 for 32-step ELF-B (Table~\ref{tab:compute}). At batch 1, ELF-B's 0.13\,s is faster than every \ours{} row (0.14--0.59\,s). At batch 32, \ours{} with $K \ge 16$ reaches 60.4--70.2 documents per second against ELF-B's 47.8 (1.26$\times$ at $K{=}16$), but $K{=}8$ remains slower: 36.7 with plain sampling and 30.8 with the states-only sampler of Table~\ref{tab:main}.

Larger batches change the comparison. ELF-B moves only from 47.8 to 49.8 documents per second between batch 32 and 256, showing limited further batching benefit in this implementation. At batch 256, \ours{} reaches 62.9 with plain $K{=}8$ (1.26$\times$ ELF-B) and 93.7 with $K{=}16$ (1.88$\times$). The evaluated states-only $K{=}8$ sampler is instead near parity (48.9 against 49.8, 0.98$\times$). Thus the throughput advantage depends on both batch size and the sampler attached to the quality result.

\paragraph{Sampler overhead.}
The evaluated samplers cost more than greedy decoding: at $K{=}8$, $+2$--$9\%$ at batch 1, $+2\%$ for temperature to $+22\%$ for nucleus with the ban at batch 32, and $+3.5\%$ to $+33\%$ at batch 256, mostly for the nucleus sort over the vocabulary. This token-selection overhead is separate from the prior: the states-only prior retains the same architecture and its sampling time agrees within 5\% with the prior it replaces.

\paragraph{Autoregressive and other references.}
With CUDA graphs on both, the states-only $K{=}8$ sampler of Table~\ref{tab:main} is 6.5$\times$, 6.7$\times$ and 6.6$\times$ faster per document than the same-data AR at batch 1, 32 and 256. The AR takes 2927, 219.2 and 135.1\,ms respectively (eager on the same machine: 8141\,ms at batch 1 and 300.4\,ms at batch 32). Against GPT-2 small, batch-32 speedups range from 4.0$\times$ for states-only $K{=}8$ to 9.0$\times$ for $K{=}32$. \ours{}-L's states-only $K{=}8$ sampler reaches 22.2 documents per second at batch 32, 1.8$\times$ ELF-M and 7.9$\times$ ELF-L at 32 steps. BD3-LM and MDLM have no CUDA-graph sampler; their wall-clock entries describe the available eager implementations. FLOPs describe computation with less dependence on these implementation differences.

\section{Additional evaluations}
\label{app:robust}

We first test how the perplexity comparisons depend on the judge, entropy anchor and document length. We then examine distributional similarity, diversity, repetition and lexical persistence, before evaluating coherence and prompted continuation. These checks establish the scope of the perplexity results and the document-quality limitations summarized in Section~\ref{sec:limitations}. Unless otherwise specified, evaluations use 1000 documents and GPT-2 Large; each table specifies whether scores are measured or entropy-matched.

\subsection{Other judges and other anchor entropies}
\label{app:judge2}
\label{app:entropies}

Changing the judge preserves the ordering of the $K\le16$ rows relative to the diffusion baselines in Table~\ref{tab:judge2}. We re-scored the evaluation dumps of Table~\ref{tab:main}, its sweeps and every baseline with Qwen2.5-1.5B and Qwen2.5-7B, using each judge's own tokens for both perplexity and per-document entropy. These judges differ from GPT-2 in model family, training corpus and tokenizer. The dumps are unchanged except for the same-data AR, re-sampled at nucleus 0.86--0.88 to bracket entropy 5.16. ELF-B's 32-step entropy is 5.158 in GPT-2 tokens and 5.159 in Qwen tokens, allowing the same rounded anchor within each judge's protocol.

The broader ordering is more stable than the close comparisons. The plain $K{=}8$ margin over ELF-B shrinks from about $5\%$ to $3\%$, while \ours{}-L's states-only rows move ahead of the same-data AR under both Qwen judges and $K{=}64$ falls behind BD3-LM block~4. GPT-2 small retains its low perplexity under Qwen as well; its measured entropy is marked where it differs from 5.16. Table~\ref{tab:judge2} retains the full scores and these exceptions.

The entropy sweep identifies a separate limit: the advantage over ELF-M and ELF-L is specific to the 5.0--5.2 nat band. Across the four GPT-2 Large columns, the block-length ordering and the comparison with BD3-LM hold wherever both systems reach the target entropy; GPT-2 small remains ahead on the reported comparisons, with a roughly constant log-perplexity gap. Our perplexity rises steeply with entropy (slope: Appendix~\ref{app:metrics}), so larger ELF models overtake our plain samplers at their higher-entropy operating points. ELF-M reaches 21.65 at entropy 5.184, below \ours{}'s 22.75 at 5.164; ELF-L reaches 23.65 at 5.281, below \ours{}'s 27.64 at 5.237 and \ours{}-L's 24.90 at 5.278 (Figure~\ref{fig:sampling}(a,b)). At 32 steps, ELF-M and ELF-L read 26.3 at 5.25 and 30.3 at 5.35. Thus changing the anchor preserves some comparisons while exposing the operating range of others.

\begin{table}[!htbp]
\caption{\textbf{Generative perplexity under three judges, at four entropies and on a 512-token window.} GPT-2 Large and Qwen columns: log-linear interpolation in each judge's own tokens; window: first 512 GPT-2 Large tokens, matched at ELF-B's 32-step window entropy 4.845 (Appendix~\ref{app:length}). Values in parentheses are measured at the entropy given; ``extr.'' = window bracket extrapolated by at most 0.06 nats; ``---'' = not computed or not reached without extrapolation; $p$ = nucleus threshold; indented rows use the states-only prior of the $K{=}8$ model above them; other rows of ours use the plain sampler.}
\label{tab:judge2}
\label{tab:entropies}
\label{tab:window}
\begin{center}
\footnotesize\setlength{\tabcolsep}{3pt}
\def\cEnt#1{{\scriptsize(#1)}}
\begin{tabular}{@{}lrrrrrrr@{}}
\toprule
& \multicolumn{4}{c}{GPT-2 Large, at entropy} & \multicolumn{2}{c}{Qwen2.5, at 5.16} & Window \\
\cmidrule(lr){2-5}\cmidrule(lr){6-7}
System & 5.00 & 5.10 & 5.16 & 5.25 & 1.5B & 7B & at 4.845 \\
\midrule
\ours{} $K{=}64$ & 28.5 & 34.5 & 39.2 & 48.4 & 35.0 & 32.3 & 38.1 \\
\ours{} $K{=}32$ & 23.4 & 28.3 & 32.2 & 40.6 & 28.0 & 25.7 & 26.7 \cEnt{extr.} \\
\ours{} $K{=}16$ & 17.7 & 21.8 & 24.9 & 31.4 & 22.0 & 20.2 & 22.2 \\
\ours{} $K{=}8$ & 15.9 & 19.7 & 22.8 & 27.64 \cEnt{5.237} & 20.1 & 18.4 & 20.4 \\
\quad states-only, $p{=}0.6$ & --- & --- & 19.4 & --- & 16.7 & 15.1 & 17.1 \\
\quad states-only, $p{=}0.55$, ban & --- & --- & 19.0 & --- & 16.5 & 14.8 & 17.0 \\
\midrule
\ours{}-L $K{=}16$ & --- & 17.9 & 20.3 & 25.2 & 18.0 & 16.4 & 18.2 \\
\ours{}-L $K{=}8$ & 13.0 & 16.0 & 18.4 & 23.2 & 16.9 & 15.5 & 16.9 \\
\quad states-only, $p{=}0.6$ & --- & --- & 16.0 \cEnt{5.144} & --- & 14.9 & 13.4 & 15.1 \\
\quad states-only, $p{=}0.5/0.55$, ban & --- & --- & 16.4 & --- & 14.8 & 13.1 & 15.0 \\
\midrule
ELF-B, 32 steps & --- & --- & 24.0 \cEnt{5.158} & --- & 20.7 \cEnt{5.159} & 19.0 \cEnt{5.159} & 25.8 \cEnt{anchor} \\
ELF-B, 64 steps & --- & --- & 19.4 \cEnt{5.096} & --- & 16.4 \cEnt{5.097} & 15.0 \cEnt{5.097} & --- \\
ELF-M, 64 steps & --- & --- & 21.7 \cEnt{5.184} & --- & 19.5 \cEnt{5.189} & 17.9 \cEnt{5.189} & 20.8 \cEnt{extr.} \\
ELF-L, 64 steps & --- & --- & 23.7 \cEnt{5.281} & --- & 20.3 \cEnt{5.265} & 18.3 \cEnt{5.265} & 16.3 \cEnt{extr.} \\
BD3-LM, block 4 & --- & 31.5 & 42.1 & 66.6 & 33 & 31 & 22.0 \\
BD3-LM, block 8 & --- & 38.8 & 52.4 & 94.8 & 41 & 39 & 28.2 \cEnt{extr.} \\
BD3-LM, block 16 & --- & 40.3 & 58.8 & --- & 41 & 39 & 24.2 \cEnt{extr.} \\
MDLM, $p{=}0.9$ & --- & --- & 35.4 \cEnt{4.90} & --- & 42.5 \cEnt{4.98} & 40.3 \cEnt{4.98} & --- \\
AR, same data, nucleus & --- & --- & 16.0 & --- & 16.0 & 15.3 & 15.0 \\
GPT-2 small & --- & --- & 13.2 & 16.3 & 13.2 & 13.2 \cEnt{5.19} & 12.7 \\
Held-out OpenWebText & --- & --- & 16.9 \cEnt{5.41} & --- & 14.0 \cEnt{5.40} & 10.8 \cEnt{5.40} & --- \\
\bottomrule
\end{tabular}
\end{center}
\end{table}

\subsection{Document length, MAUVE and a length-controlled window}
\label{app:length}
\label{app:mauve}

Document length matters because per-document entropy grows with length: shorter outputs start lower on the entropy axis and are moved further up their perplexity--entropy curve when matched to 5.16. Although the canvas is fixed at $L=1024$, scoring stops at the first end-of-text token. Held-out documents average 956 T5 tokens and 65\% end before the canvas; ours average 959--979 tokens. ELF-B and ELF-M usually fill the canvas (1016--1018 tokens, 98--99\% at the canvas), whereas BD3-LM and MDLM are shorter (means 752 and 733; medians 872 and 828, against 985 for held-out text). We therefore evaluate a common 512-judge-token window for both re-scoring and MAUVE; Cosmos uses a separate MAUVE@256 protocol (Appendix~\ref{app:cosmos}).

The window check preserves the comparison with ELF-B and substantially narrows the gap to BD3-LM. Table~\ref{tab:window} re-scores both perplexity and entropy on the first 512 GPT-2 Large tokens and re-matches at ELF-B's 32-step window entropy, 4.845. The plain $K{=}8$ and states-only samplers are respectively $21\%$ and $34\%$ ahead of ELF-B (20.4 and 17.0--17.1 against 25.8). Their advantage over BD3-LM block~4 falls from roughly a factor of two to 7--23\%: its window score is 22.0, bracketed by fresh samplings at nucleus 0.80 and 0.85. The same-data AR and GPT-2 small remain ahead (15.0 and 12.7). This all-documents window includes outputs shorter than 512 tokens; restricting to longer documents moves our rows by at most 0.2 but retains only 68\% of BD3-LM's documents. Thus part of the original BD3-LM gap comes from the length protocol.

MAUVE reveals a distributional cost of the lowest-perplexity settings. At $K\le16$, the plain samplers score 0.90--0.93, numerically matching or exceeding every evaluated diffusion baseline and GPT-2 small, while the same-data AR reaches 0.93--0.95. The states-only scores fall to 0.85--0.89 for \ours{} and 0.88--0.91 for \ours{}-L, compared with 0.90--0.93 for their plain samplers. These ranges are above ELF-B at 32 steps (0.73), around its 64-step score (0.89), and below the same-data AR; plain $K{=}64$ is also lower at 0.87. The standard deviation over five quantization seeds is 0.003--0.037. MAUVE measures similarity to the held-out distribution in the judge's feature space (Appendix~\ref{app:metrics}); the decrease motivates the diversity checks below, while document coherence is evaluated separately.

\subsection{Diversity and overlap with the training set}
\label{app:memo}
\label{app:position}

Per-document entropy can remain high even when generated documents resemble one another. Table~\ref{tab:memo} therefore measures diversity across the dump: \emph{Distinct-4} is the fraction of distinct T5-token 4-grams, and \emph{Self-BLEU-4}, computed on lowercased words, scores each of 300 documents against the other 299 (higher means more resemblance). Every generator is less diverse than real text by these measures. Our plain rows resemble ELF-B (Self-BLEU 0.301 and 0.296 against 0.292), while the states-only operating points are less diverse (0.36--0.40, with distinct-4 0.74--0.75); all our rows have higher Self-BLEU than the same-data AR (0.23). The lower-perplexity configurations thus also concentrate the generated document distribution.

Verbatim training overlap addresses a different question. We index 13-grams from the first 1024 T5 tokens of two million training documents (1.9B tokens), then measure the fraction of each evaluation dump's 13-grams found in the index. Held-out documents were never indexed and provide a reference for overlap within OpenWebText. The states-only rows share 0.5--1.0\% of their 13-grams with the index, below GPT-2 small's 1.4\%; autoregressive decoders generally overlap more than token-level diffusion models (0.4--1.4\% against 0.04--0.15\%), with \ours{}-L highest among ours. All measured overlap rates remain below held-out text (2.7\%). This statistic measures overlap with the indexed training subset, including shared quotations and boilerplate; it is not a measure of text quality.

A separate position check finds no late increase in judge negative log-likelihood: after pooling tokens into 128-token bins, the ratio of the second half to the second bin is 0.82--0.91 for our models, against 0.85 for held-out text and 0.90 for ELF-B. This local predictability diagnostic leaves open whether the document remains coherent, which we examine in Appendix~\ref{app:coherence}.

\begin{table}[!htbp]
\caption{\textbf{Cross-sample diversity and 13-gram overlap with a 2M-document training index.} Self-BLEU uses 300 documents; overlap is the fraction of each dump's T5-token 13-grams found in the index.}
\label{tab:memo}
\begin{center}
\footnotesize\setlength{\tabcolsep}{6pt}
\begin{tabular}{lrrr}
\toprule
System & Distinct-4 $\uparrow$ & Self-BLEU-4 $\downarrow$ & 13-gram overlap \\
\midrule
Held-out OpenWebText & 0.905 & 0.148 & 2.67\% \\
\ours{} $K{=}8$, $t{=}0.75$ & 0.803 & 0.301 & 0.54\% \\
\ours{}-L $K{=}8$, $t{=}0.70$ & 0.781 & 0.296 & 0.83\% \\
\ours{} $K{=}8$, states-only, nucleus 0.55, ban & 0.750 & 0.404 & 0.49\% \\
\ours{}-L $K{=}8$, states-only, nucleus 0.6 & 0.741 & 0.363 & 0.96\% \\
ELF-B, 32 steps & 0.801 & 0.292 & 0.04\% \\
BD3-LM, block 4, nucleus 0.9 & 0.898 & 0.207 & 0.15\% \\
MDLM, nucleus 0.9 & 0.859 & 0.195 & 0.04\% \\
GPT-2 small, $t{=}1$ & 0.810 & 0.270 & 1.38\% \\
AR, same data, $t{=}0.88$ & 0.837 & 0.231 & 0.42\% \\
\bottomrule
\end{tabular}
\end{center}
\end{table}

\subsection{Repetition and the 4-gram ban}
\label{app:norepeat}

The measured repetition can be suppressed at sampling time with little change in matched-entropy perplexity. Without a ban, our $K\le16$ loop rates are 7--12\%, above MDLM (0.3\%), BD3-LM (0.6--1.7\%) and ELF-B at 32 steps (4.0\%; 14.6\% at 64 steps). Because the $K$ blocks decode position-synchronously, at step $s$ each block can inspect all tokens emitted at earlier steps. A token completing an $n$-gram already present anywhere in the document receives logit $-\infty$ before the temperature or nucleus cut, adding one tensor comparison per step.

At fixed temperature, the ban raises entropy by 0.05--0.10 nats and perplexity rises with it. Table~\ref{tab:norepeat} instead compares each banned run with the unbanned sweep at the same entropy: changes are $-2\%$ to $+1\%$ for $n\ge4$, comparable to the observed seed variation, and $-4.5\%$ for $n=3$. At entropy 5.16, the 4-gram ban gives 22.7 against 22.8 for \ours{} $K=8$ and 19.0 against 19.4 with its states-only prior.\footnote{The other anchor readings are 25.3 against 24.9 for \ours{} $K=16$ and 17.6 against 17.6 for \ours{}-L $K=8$ (nucleus; extrapolated 0.03 nats). The large states-only model reads 16.4 with the ban at 5.16; its unbanned sweep reaches 16.0 at entropy 5.144.} No banned run triggers the loop criterion, an 8-gram occurring at least three times in the first 256 T5 tokens (Appendix~\ref{app:metrics}). This demonstrates control of that repetition pattern; broader lexical and discourse consistency require the checks below. The \ours{} states-only row of Table~\ref{tab:main} uses the ban, whereas its \ours{}-L and baseline rows do not.

\begin{table}[!htbp]
\caption{\textbf{Document-wide $n$-gram ban at sampling time.} ``PPL without ban'' is the unbanned sweep interpolated at the banned run's entropy; ``Loop, no ban'' is the unbanned model's loop rate at the nearest setting (``---'' = not measured). Every banned run has a loop rate of 0\%.}
\label{tab:norepeat}
\begin{center}
\footnotesize\setlength{\tabcolsep}{4pt}
\begin{tabular}{lcrrrrr}
\toprule
Sampler & $n$ & Entropy & PPL with ban & PPL without ban & $\Delta$ PPL (\%) & Loop, no ban \\
\midrule
\multicolumn{7}{l}{\emph{\ours{} $K{=}8$, prior's codes}} \\
$t{=}0.65$ & 4 & 5.111 & 20.25 & 20.2 & $0$ & 9\% \\
$t{=}0.70$ & 4 & 5.162 & 22.76 & 22.9 & $-0.7$ & 6.9\% \\
$t{=}0.70$ & 3 & 5.235 & 26.25 & 27.5 & $-4.5$ & --- \\
$t{=}0.70$ & 6 & 5.120 & 20.50 & 20.6 & $-0.5$ & --- \\
$t{=}0.70$ & 8 & 5.109 & 19.82 & 20.1 & $-1.4$ & --- \\
nucleus 0.55 & 4 & 5.170 & 21.87 & 21.8 & $+0.5$ & 7\% \\
\addlinespace[2pt]
\multicolumn{7}{l}{\emph{\ours{} $K{=}8$, states-only prior}} \\
nucleus 0.55 & 4 & 5.158 & 18.95 & 19.3 & $-2$ & 7.4\% \\
\addlinespace[2pt]
\multicolumn{7}{l}{\emph{\ours{} $K{=}16$, prior's codes}} \\
$t{=}0.60$ & 4 & 5.120 & 23.01 & 22.8 & $+0.9$ & 9\% \\
$t{=}0.65$ & 4 & 5.166 & 25.34 & 25.3 & $+0.3$ & 7.6\% \\
\midrule
\multicolumn{7}{l}{\emph{\ours{}-L $K{=}8$, prior's codes}} \\
$t{=}0.70$ & 4 & 5.228 & 21.52 & 21.9 & $-1.8$ & 8\% \\
nucleus 0.50 & 4 & 5.190 & 18.86 & 18.8 & $+0.3$ & 10\% \\
nucleus 0.55 & 4 & 5.226 & 20.51 & 20.6 & $-0.4$ & 8\% \\
\addlinespace[2pt]
\multicolumn{7}{l}{\emph{\ours{}-L $K{=}8$, states-only prior}} \\
nucleus 0.50 & 4 & 5.123 & 15.32 & 15.3 & $+0.4$ & 10--16\% \\
\bottomrule
\end{tabular}
\end{center}
\end{table}

\paragraph{The same ban on the autoregressive baselines.} Applying the ban to both sides preserves an AR advantage. After re-sampling at lower thresholds and matching entropy 5.16, the banned same-data AR and GPT-2 small read 15.2 and 12.4 against 16.0 and 13.2 without the ban: a 5--6\% improvement, with no measured loops. The banned states-only rows of \ours{} (19.0) and \ours{}-L (16.4) are then 25\% and 8\% behind the same-data AR, instead of 19\% and 3\% against its unbanned sampler (21\% for our unbanned states-only row, Appendix~\ref{app:ar}).

\emph{AR sampling details.} At unchanged settings, the ban raises entropy by 0.10--0.15 nats: same-data AR moves from 15.4 at 5.151 (nucleus 0.88) to 19.7 at 5.258, and GPT-2 small from 12.1 at 5.122 ($t=1$) to 15.9 at 5.271. The new brackets are nucleus 0.80--0.85 for same-data AR (13.65 at 5.125; 16.80 at 5.211) and $t=0.9,1.0$ for GPT-2 small (12.30 at 5.157; 15.88 at 5.271).

\subsection{Topic persistence and lexical coupling}
\label{app:topic}

\begin{figure}[!htbp]
\centering
\includegraphics[width=\linewidth]{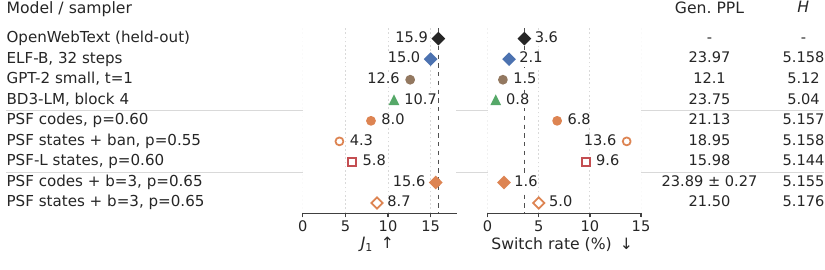}
\caption{\textbf{Lexical topic-persistence measures.} $J_1$ compares adjacent-chunk
word overlap with cross-document overlap; switch rate counts low-overlap transitions.
All our rows use $K=8$.
$p$: nucleus threshold; states: states-only prior; ban: 4-gram ban; $b$: lexical
bonus. Dashed lines mark held-out text. Right: GPT-2 Large perplexity and entropy
$H$; error bar / $\pm$: three-seed standard deviations.}
\label{fig:topic}
\end{figure}

Lower perplexity in the states-only configurations is accompanied by weaker lexical persistence across chunks. Each block samples its realization of the shared plan, with limited coupling through the landing target. Figure~\ref{fig:topic} shows the resulting lexical continuity: our one-shot nucleus sample has $J_1=8.0$ and a switch rate of 6.8\%, while the states-only rows have $J_1=4.3$--5.8 and switch rates of 10--14\%. Real OpenWebText, ELF-B and GPT-2 small have higher overlap ($J_1$ 13--16) and fewer switches (1.5--3.6\%). BD3-LM illustrates that the two measures need not rank systems identically: its $J_1$ is lower at 10.7 but its switch rate is only 0.8\%.

For these measures, we split documents into consecutive chunks of 45 content words after removing stop words, punctuation and numbers. $J_1$ is the mean Jaccard overlap of adjacent chunks' word sets divided by the mean overlap between chunks from different documents; the \emph{switch rate} counts adjacent pairs within 1.5 times this cross-document floor. To test whether directly sharing lexical choices helps, we add a sampling-time bonus: after the nucleus cut, tokens with T5 identifiers $\ge2000$ that another block has emitted receive logit bonus $b$. This token-ID rule is a heuristic for selecting content tokens.

With $b=3$, the code-augmented $K=8$ model reaches $J_1=15.6\pm0.4$ and a 1.6\% switch rate over three seeds, at perplexity $23.89\pm0.27$ and entropy 5.155. The gain costs 10--14\% perplexity at equal entropy relative to the uncoupled nucleus sweep.\footnote{At nucleus 0.60, the coupled model reads 20.8 at entropy 5.107 against 18.9 without coupling ($+10\%$); at nucleus 0.65 it reads 23.89 at 5.155 against 21.0 ($+14\%$).} With the states-only prior, the bonus recovers only the lexical persistence of the uncoupled code-augmented row ($J_1=8.7$, switch rate 5.0\%), at a 6\% perplexity cost at equal entropy (20.6 against 19.4). This partial recovery is consistent with the codes helping coordinate lexical realization. Because the bonus directly rewards word reuse, improved overlap alone does not establish discourse coherence; the next evaluation tests that separately. The coupled points remain outside Table~\ref{tab:main} because they trade perplexity for lexical persistence.

\subsection{LLM-judged coherence}
\label{app:coherence}

The perplexity margins do not carry over to the coherence judgements tested here. We use two instruction-tuned judges in blinded pairwise comparisons, drawing 200 seeded document pairs for each system pair and presenting each in \emph{both} orders. Only pairs whose preferred passage is unchanged by the order count as consistent. Qwen2.5-7B-Instruct (bf16, greedy) reads the first 300 words and judges internal coherence (one topic, consistent facts, no contradictions) and resemblance to a single human-written article. Qwen2.5-32B-Instruct judges coherence on words 1--300, words 300--600 (past the first two or three block seams), and the whole document. The real-against-real row calibrates the 7B judge, which has a strong slot preference.

Both judges strongly prefer real text to every generator they evaluate (Table~\ref{tab:coherence}). Real text wins 96--99\% of the 7B judge's consistent pairs against our $K=8$ configurations, compared with about 93\% against ELF-B, 90\% against BD3-LM and 88\% against GPT-2 small. Under the 32B judge the corresponding preferences are 99--100\% for every tested generator, with fewer order-inconsistent pairs in these real-text comparisons. The table retains all win counts so that these conditional percentages can be read with their denominators.

Direct generator comparisons also limit the perplexity claim. The small states-only + ban configuration loses to ELF-B, BD3-LM and the same-data AR in each tested window. Under the 32B judge, the plain sampler loses to ELF-B on the first 300 words, with a smaller imbalance in the middle window. Its whole-document result is inconclusive: the 10:13 count contains only 23 of the 200 pairs, with 177 excluded for order inconsistency, so it does not establish parity. The states-only + ban count is 7:28. The middle-window deficit is no larger. Order-inconsistent pairs indicate unstable judgements and cannot be assumed to be near-ties.

\begin{table}[!htbp]
\caption{\textbf{Blinded pairwise LLM judgement,} 200 document pairs per row, each shown in both slot orders. Cells: consistent pairs won by system 1 : won by system 2; the remaining pairs disagree between orders. The small ``states-only + ban'' configuration is the states-only \ours{} row of Table~\ref{tab:main}; the \ours{}-L configuration here also uses the ban, unlike its main-table row.}
\label{tab:coherence}
\label{tab:coherence32}
\begin{center}
\footnotesize\setlength{\tabcolsep}{4pt}
\begin{tabular}{@{}lccccc@{}}
\toprule
& \multicolumn{2}{c}{Qwen2.5-7B, words 1--300} & \multicolumn{3}{c}{Qwen2.5-32B coherence, words} \\
\cmidrule(lr){2-3}\cmidrule(l){4-6}
System 1 vs.\ system 2 & Coherent & Human-written & 1--300 & 300--600 & all \\
\midrule
real vs.\ real (calibration) & 65:42 & 41:41 & --- & --- & --- \\
\ours{} plain (prior's codes) vs.\ real & 6:143 & 12:131 & 1:198 & 0:195 & 1:196 \\
\ours{} states-only + ban vs.\ real & 1:166 & 2:147 & 0:194 & 0:197 & 0:196 \\
\ours{}-L states-only + ban vs.\ real & 4:144 & 10:147 & --- & --- & --- \\
ELF-B, 32 steps vs.\ real & 10:127 & 7:121 & 0:193 & 0:194 & 1:196 \\
BD3-LM, block 4 vs.\ real & 13:117 & 15:94 & 2:189 & 1:191 & 1:195 \\
GPT-2 small, $t{=}1$ vs.\ real & 14:102 & 18:105 & --- & --- & --- \\
AR, same data vs.\ real & --- & --- & 2:177 & 2:176 & 2:183 \\
\midrule
\ours{} plain vs.\ ELF-B 32 & --- & --- & 19:48 & 29:37 & 10:13 \\
\ours{} states-only + ban vs.\ ELF-B 32 & 13:43 & 38:78 & 18:46 & 21:39 & 7:28 \\
\ours{} states-only + ban vs.\ BD3-LM 4 & 16:73 & 22:102 & 13:102 & 9:83 & 2:93 \\
\ours{} states-only + ban vs.\ AR, same data & --- & --- & 4:94 & 7:102 & 1:62 \\
\bottomrule
\end{tabular}
\end{center}
\end{table}

\subsection{Prompted continuation without retraining}
\label{app:prompted}

Predictive states allow prompted continuation through plan inpainting without retraining for this task. Given the first $P=256$ tokens of a held-out document ($m=P/\ell$ blocks), the encoder yields the exact states $\vs_1,\dots,\vs_{m+1}$ and codes $\vr_1,\dots,\vr_m$. The prior fills the remaining plan by replacement inpainting: at every Euler step, the given entries are reset to their exact noisy value, RePaint-style, optionally with $r$ resampling passes per step. All blocks then decode in parallel, with prompt blocks forced to the given tokens.

We evaluate 500 held-out documents with at least 512 tokens, measuring continuation perplexity under GPT-2 Large given the prompt, per-document entropy, and \emph{prompt relevance}. Relevance is the fraction of continuation content words (alphabetic, at least four letters, not a stop word) occurring in the prompt, minus the fraction for mismatched prompt--continuation pairs (chance, about 2\%). GPT-2 small and the same-data AR continue the same prompts. The ``oracle plan'' uses the encoder's plan of the full real document with only the prompt blocks forced, providing a decoder reference with the prior bypassed.

\begin{figure}[!htbp]
\centering
\includegraphics[width=\linewidth]{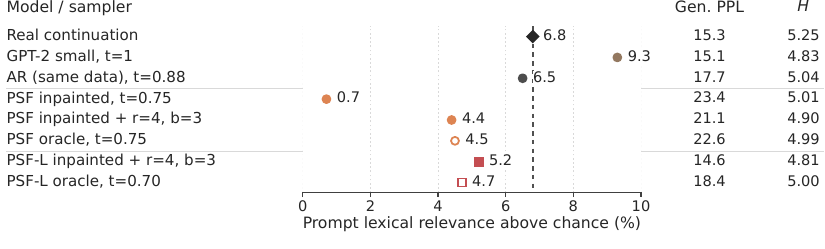}
\caption{\textbf{Prompted continuation} of 500 held-out documents from 256-token
prompts. Relevance is content-word overlap above chance. Our filled points use
inpainted plans; hollow points use oracle plans; all use $K=8$. $r$: resampling
passes; $b$: lexical bonus. Dashed: real continuation. Right: continuation
perplexity and entropy $H$ under GPT-2 Large.}
\label{fig:prompted}
\end{figure}

Inpainting alone gives weak lexical relevance: 0.7\% above chance with the plain sampler. Combining $r=4$ resampling passes and lexical coupling $b=3$ raises it to 4.4--5.2\%, compared with 6.8\% for real continuations and 6.5\% for the same-data AR (Figure~\ref{fig:prompted}). The score reaches the range of the measured oracle-plan rows (4.5--4.7\%), but those rows use different samplers and entropies, so they do not establish an upper bound on prompt use. Perplexities span 14.6--23.4, against 15.1 for GPT-2 small and 17.7 for the same-data AR, also at different entropies. GPT-2 small's overlap exceeds that of real continuations, illustrating that lexical reuse alone does not establish continuation quality.

\paragraph{A separately fine-tuned prompt-attention path.} To test direct access to prompt tokens, we add a third sub-layer per block that cross-attends to them (19M parameters) and fine-tune for 6k steps. Unconditional perplexity remains similar (22.4 at entropy 5.16), while raw prompt overlap is 3.2\% against 3.0\% with the prompt withheld. Under teacher forcing with true predictive states, the extra prompt tokens lower continuation negative log-likelihood by only 0.2\%, suggesting little training incentive to use this path in that setup. This negative result motivates testing a prompt-conditioned prior or training under corrupted predictive states.

\FloatBarrier
\section{Selected samples}
\label{app:samples}

These five documents come from 1000-document evaluation dumps using the prior's codes. Following ELF's appendix format, each box names the model and sampler, gives the document's unigram entropy (nats, GPT-2 tokens) and GPT-2 Large perplexity, and displays up to its first 1300 characters unedited, cut at a word boundary, with ``\ldots'' marking truncation. Scores use up to the first 1024 judge tokens of each document.

\begin{samplebox}{\ours{} $K{=}8$ OWT, plain sampler ($t{=}0.75$)}
\ttfamily\footnotesize\raggedright
Entropy: 5.24 \quad Gen. PPL: 18.47\par\medskip
"We are tackling a number of challenges with a variety of species," he told the BBC. "Although we're trying to expand our knowledge of the ecology of eucalyptus, we also need to develop solutions to address these environmental issues. We're also looking to develop an ethical approach to enhance our knowledge of the eucalyptus flora that are essential for the food chain." The researchers are using labs to develop biogeochems to assess the impact of global forest ecologies on the plant and the way "plants and animals and soil composition," said Nelson. "The idea is that this is what we're seeing now -- the ability to filter our groundwater for biofuels and other biological materials. We know that we're entering a world where organic fertilizers and their effects on water supplies are already affecting marine organisms and the ocean." Agri-giant bacteria will need to adapt quickly and adapt to new circumstances, such as the proliferation of bacteria and viruses. With new technologies and methods, ecosystems will be able to use insecticides and other herbicides to control insects' growth. But for now, ecology will have a significant impact on the global food supply. "Today's announcement is a significant step toward a world that is on track to do much more to prevent further impacts...
\end{samplebox}

\begin{samplebox}{\ours{} $K{=}8$ OWT, nucleus 0.62}
\ttfamily\footnotesize\raggedright
Entropy: 5.23 \quad Gen. PPL: 14.96\par\medskip
But the EPA's latest report is not a conclusion that will be adopted anytime soon. Instead, it is just a few paragraphs from a statement from the White House. As of January 1, the U.S. Environmental Protection Agency estimates that the EPA's proposed carbon-reduction goal of reducing carbon dioxide emissions by 80 percent by 2030 is about 30 percent lower than the current projections. "I don't think the projections are out of date," the report says. "I think the big estimate of the costs of new coal-fired power plants -- including total emissions of carbon-- is already more than double the level of subsidies that have been available to utilities since the mid-1990s. "It's very difficult to predict the actual benefits," said Jason Lowenthal, a climate change economist at the nonpartisan Institute for Energy Economics and Financial Analysis. "It is hard to quantify the benefits, but you can estimate the costs of a transition to a low-carbon future." The government noted that "the increase in the emission of greenhouse gases from 1970 to present, by an average of five cents per metric ton, would have resulted in additional reductions." But, of course, carbon emissions are not the only source of greenhouse gas emissions. It's that simple. In fact, some estimates put the total amount...
\end{samplebox}

\begin{samplebox}{\ours{}-L $K{=}8$ OWT, nucleus 0.55}
\ttfamily\footnotesize\raggedright
Entropy: 5.16 \quad Gen. PPL: 9.91\par\medskip
I believe the American president should have a backup plan, but he has a point. And if he had, he would have had to do it. When I think about what happened to my father in the last year of his presidency, I think about what he'd have to do to prevent the end of his presidency. But I think the outcome of the election is a consequence of our nation's history: When I was in the midst of a deep recession, I suffered from a traumatic episode that included both high-profile and low-level violence. Against my father, I didn't think I would ever have a chance to be president again. I was a man of the people, and I was wrong about a lot of things. But the United States's wars have been a part of our national security strategy for decades, and it's hard to believe that our military was a major part of the solution to our long-term problems. When Barack Obama was elected, he wisely embraced the Republican position, which was to end all wars at once, beginning with the first Gulf War and ending with the second. My bet, now, that's what I've been focusing on. But it's also why I'm fighting for the same kind of solutions that I believe in. We need to work together to make sure that we're delivering on our promise to get a deal that works for all Americans. We need to get our economy growing...
\end{samplebox}

\begin{samplebox}{\ours{}-L $K{=}8$ OWT, nucleus 0.60}
\ttfamily\footnotesize\raggedright
Entropy: 5.32 \quad Gen. PPL: 18.29\par\medskip
"We are committed to delivering high quality, safe and reliable air quality services to consumers and businesses. The Government believes that in order to achieve the full cost savings of the air quality improvements we are delivering, Transport Scotland will be taking steps to ensure that this cost savings are achieved," the Scottish Government said. The ASA's complaint follows calls from ministers to reduce the number of hazardous substances detected by Transport Scotland. In a statement, Transport Scotland said: "The Scottish Government believes that any unnecessary risks to road users and vehicles are minimal and should be treated as such. "In seeking to stop the improper use of these, we will also monitor the performance of other suppliers to ensure they meet the regulatory requirements of the relevant regulators and the regulations they are required to follow. "We will also look at the progress of the inspection and audit process to ensure that the safety and integrity of the automotive industry is not compromised." On Wednesday, Mr Carswell said the regulators were "fully able to respond to the regulatory and audit requirements" but he added: "These assessments are based on rigorous scientific and technological analysis, which has shown the effectiveness of some car...
\end{samplebox}

\begin{samplebox}{\ours{}-L $K{=}16$ OWT, plain sampler ($t{=}0.65$)}
\ttfamily\footnotesize\raggedright
Entropy: 5.35 \quad Gen. PPL: 19.61\par\medskip
"I think it's important to say that we're all aware of it," said the singer. "We all have different experiences. When I first started playing in the band, I just remember a flurry of rapping and slurring, but it was solid. "It's like we lost the first time. We're not going to be able to get back together again."[5] The band had a falling-out with legendary guitarist Pete Wentz in April of that year. He was upset and stated that future X-Men would be too important to the album because he was so focused on the album. But he said that he never felt like it was going to be a new record. "It was crazy. When I found out that we were going to Heaven and Hell, I never really felt like that. Everywhere you look there he does, it's a hell of a lot easier to deal with than the ones we did," he said. "We don't want to be the 'Little Man' anymore." *** What is the meaning of All That Jazz in the Lifetime? Not the past. "It was the greatest moment of my life. I'm not the only one in my family who felt that way," he told NPR's Dana Loesch in August. "When I played with a couple of my closest friends, Jason and I, for whatever reason, it's my first band, I was a very big fan of his music. His pieces, especially his solos, such as All That Jazz and 'Sleepless in the Dark', and his signature...
\end{samplebox}

\end{document}